\documentclass[lettersize,journal]{IEEEtran}
\usepackage{graphicx}
\usepackage[caption=false,font=footnotesize]{subfig}
\usepackage{multirow}
\usepackage{amsmath,amssymb,amsfonts}
\usepackage{amsthm}
\usepackage{mathrsfs}
\usepackage{xcolor}
\usepackage{textcomp}
\usepackage{booktabs}
\usepackage{algorithm}
\usepackage{algorithmicx}
\usepackage{algpseudocode}
\usepackage{listings}
\usepackage{orcidlink}
\usepackage{xurl}
\usepackage{placeins}
\usepackage{stfloats}
\usepackage{acronym}
\usepackage{siunitx}
\usepackage{bm}
\usepackage{cite}
\usepackage{hyperref}
\hypersetup{hidelinks}
\usepackage[noabbrev, capitalise]{cleveref}
\allowdisplaybreaks
\acrodef{RL}{Reinforcement Learning}
\acrodef{NN}{Neural Network}
\acrodef{DoF}{Degree of Freedom}
\acrodef{ID}{Inverse Dynamics}
\acrodef{IMU}{Inertial Measurement Unit}
\acrodef{PPO}{Proximal Policy Optimization}
\acrodef{PSD}{Power Spectral Density}
\acrodef{MLP}{Multi-Layer Perceptron}
\acrodef{COG}{Center of Gravity}
\acrodef{EE}{end-effector}
\acrodef{IK}{Inverse Kinematics}
\acrodef{AOA}{angle of arrival}
\acrodef{RF}{radio frequency}
\acrodef{SDR}{Software Defined Radio}
\acrodef{SM}{Switching Matrix}
\acrodef{CW}{Continuous Wave}
\acrodef{ISM}{Industrial, Scientific, and Medical}
\acrodef{TDM}{Time Division Multiplexing}
\acrodef{NFC}{Negative Flow Control}
\acrodef{PFC}{Positive Flow Control}
\acrodef{LS}{Load Sensing}
\acrodef{MPC}{Model Predictive Control}
\acrodef{FF}{Feed Forward}
\acrodef{ECU}{Engine Control Unit}
\acrodef{DCV}{Directional Control Valve}
\acrodef{LUT}{Lookup Table}
\acrodef{MISO}{Multiple Input Single Output}
\acrodef{MPM}{Material Point Method}
\acrodef{SPH}{Smoothed Particle Hydrodynamics}
\acrodef{DEM}{Discrete Element Method}
\acrodef{FEE}{Fundamental Earth-moving Equations}

\newcounter{validationquestion}[subsection]
\renewcommand{\thevalidationquestion}{Q\arabic{validationquestion}}

\crefformat{validationquestion}{#2#1#3}
\Crefformat{validationquestion}{#2#1#3}
\newcommand{\vq}[2]{%
  \refstepcounter{validationquestion}%
  \paragraph*{\thevalidationquestion: #2}%
  \label{#1}%
}

\begin{document}

\title{Size Doesn't Matter: Material-State Reinforcement Learning for Excavator Transferable Soil Manipulation}

\author{Lennart~Werner\orcidlink{0000-0002-1338-0458},
        Pol~Eyschen\orcidlink{0009-0001-2371-4116},
        Sean~Costello\orcidlink{0000-0002-1889-8467},
        Pierluigi~Micarelli\orcidlink{0009-0009-0016-8683},
        Andrei~Cramariuc\orcidlink{0000-0002-9301-0253},
        and~Marco~Hutter\orcidlink{0000-0002-4285-4990}%
\thanks{L. Werner, P. Eyschen, A. Cramariuc, and M. Hutter are with the Robotic Systems Lab, ETH Z\"urich, Leonhardstrasse 21, 8092 Z\"urich, Switzerland (e-mail: lennartwerner@ethz.ch; peyschen@ethz.ch; crandrei@ethz.ch; mahutter@ethz.ch).}%
\thanks{S. Costello and P. Micarelli are with Hexagon Innovation Hub GmbH, Heinrich-Wild-Strasse 201, 9435 Heerbrugg, Switzerland (e-mail: sean.costello@hexagon.com; pierluigi.micarelli@hexagon.com).}}

\markboth{Preprint}%
{Werner \MakeLowercase{\textit{et al.}}: Size Doesn't Matter: Material-State Reinforcement Learning for Excavator Soil Manipulation}

\maketitle

\begin{abstract}
Earthmoving tasks such as excavation, backfilling, or embankment construction require deliberate repositioning of deformable soil.
For these tasks, human operators use all shovel faces, while autonomous systems so far are limited to excavation and dumping.
Current methods often rely on heuristic models but do not incorporate soil mechanics.
We address this shortcoming by using Reinforcement Learning in a GPU-parallelized Material Point Method particle simulation.
Our controllers are conditioned on material state such as shape and compactness, enabling skills that use multiple contact faces of the tool and displace material both inside and outside of the shovel.
To use the same learned weights across machines, our policies operate in a normalized end-effector space and are deployed through a calibrated machine interface.
We evaluate this calibrated transfer on an \SI{11.5}{\tonne} hydraulic excavator and a \SI{500}{\gram} tabletop robot.
We validate performance through autonomous construction of a \SI{42}{\m} long, \SI{2.1}{\m} high embankment in \SI{45}{\min}, executing 201 individual policy strokes without failure, retry, or operator intervention.
In a direct comparison, the autonomous controller matches an expert operator's progression speed and produces a higher, more consistent embankment.
Additional qualitative backfilling and compaction experiments demonstrate the material-state awareness and calibrated transfer across machines.
\end{abstract}


\section{Introduction}
\IEEEPARstart{T}{he} large majority of heavy construction machinery is operated by trained human drivers.
This job is often hazardous and physically demanding, while at the same time, the construction sector faces a persistent shortage of experienced operators~\cite{dadhich2016earthmoving}.
Automation can create safer work sites while simultaneously improving productivity and precision.

Earthmoving requires two fundamental skills: 
accurate machine motion control and understanding of soil mechanics that allows the tool to manipulate material effectively.
We take inspiration from the human ability to transfer soil-manipulation skills between machines of different sizes and actuation without relearning how a tool interacts with the ground.
Our method separates the machine-specific motion control from the transferable deformable soil manipulation.
We learn the latter as a single-stroke, end-effector space interaction controller that operates in the arm plane of the excavator.
Our controllers leverage soil mechanics for task completion.
For excavators, depending on soil conditions, changes in insertion depth, bucket angle, or contact side determine whether soil is cut, pushed, lifted, spilled, compacted, or allowed to collapse.
This makes automated soil reshaping a challenging soft-body manipulation problem.
Existing methods commonly use simplified soil-interaction models or geometric terrain updates. 
These support dig-and-dump cycles, where a planned filling path and target dumping location approximate the desired outcome, but they are less suitable when terrain shape and material flow are a vital element of the interaction.
Forming an embankment in one stroke, for example, requires exploiting soil flow around the cutting edge and in front of the bucket.
Such contact-rich interactions are difficult to achieve with hand-designed trajectories.
In this paper, we investigate the following contact-rich tasks:
\begin{itemize}
    \item \emph{Embankment formation}: a single stroke creates a negative trench region and a positive elongated embankment by exploiting the flow of displaced material inside and out of the bucket.
    \item \emph{Backfilling}: the shovel pushes material into a depression and can use multiple contact faces of the bucket.
    \item \emph{Compaction}: the shovel applies pressure to increase the compaction state of a soil structure while avoiding excessive forces that would collapse or displace the material.
\end{itemize}
We evaluate embankment formation quantitatively on a full-size Menzi Muck M445 hydraulic excavator (M445) and use a small tabletop robot arm (LeExcavator) to demonstrate calibrated transfer of the same policy checkpoint at tabletop scale.
Backfilling and compaction are evaluated as qualitative demonstrations of the same formulation on both platforms.

\begin{figure}
\centering
\includegraphics[width=\linewidth]{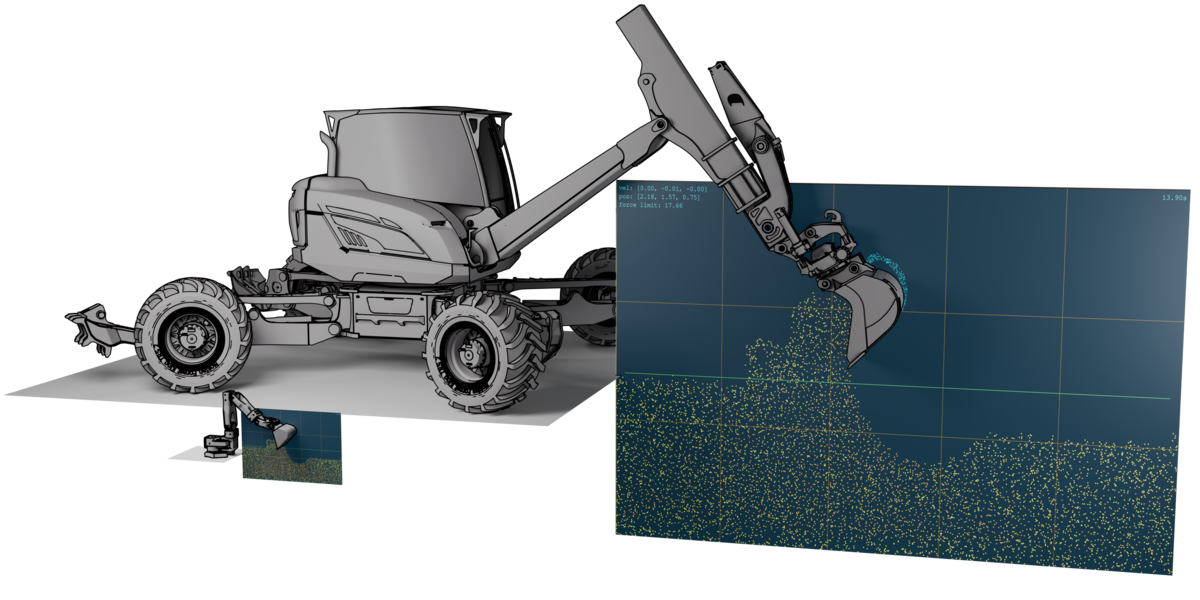}
\caption{Overview of the proposed setup: local two-dimensional control policy trained in a \acs{MPM} particle simulation executed on the \SI{11.5}{\tonne} M445 hydraulic excavator and \SI{500}{\gram} tabletop robot through a low-level end effector space motion control interface.}
\label{figOpen}
\end{figure}

The main technical contribution is the simulator-policy formulation.
We use a custom, parallelized, two-dimensional \acf{MPM} particle simulation as shown in Figure~\ref{figOpen} with a prescribed shovel body, grid-based shovel-soil contact, and a compaction model.
The simulated soil state can be directly used in task rewards, which enables earthmoving objectives that are only weakly observable from surface height, such as surface hardness.
Larger worksite tasks embed the learned single-stroke manipulation controller in a state machine based planner.

Our contributions are:
\begin{itemize}
    \item An excavation-oriented 2D \ac{MPM} simulator that is fast enough to train using \ac{RL} and detailed enough to expose the relevant terrain state.
    \item A hierarchical formulation that confines learning to a local skill in a normalized task space, while workspace planning and shovel-pose tracking remain conventional, so machine scale becomes a matter of interface calibration with no retraining required.
    \item A field evaluation of embankment formation, backfilling, and compaction that runs one policy checkpoint on both the \SI{11.5}{\tonne} M445 hydraulic excavator and the \SI{500}{\gram} tabletop LeExcavator arm, and compares autonomous embankment construction against an expert operator.
\end{itemize}

\section{State of the Art}
\label{sec:state_of_the_art}
Autonomous earthworks requires solving the problem of manipulating deformable, history-dependent granular material.
In literature, this task is often approached by trajectory generation and tracking, force-constrained motion planning, or using machine-specific end-to-end RL on a heuristic soil model.
For tasks such as bucket filling and dumping, these approaches have been effective, while tasks that require accounting for soil mechanics remain unsolved.
In particular, contact-rich tasks such as surface compaction, piling, or backfilling by pushing material back into the hole require an advanced understanding of tool-soil interaction.

\subsection{Model-Based Excavation Planning and Soil-Tool Interaction}
\label{subsec:sota_model_based}

Conventional trajectory optimization for excavation searches over spline and waypoint representations under geometric, swept-volume, and kinematic constraints \cite{yang2020trajectory}.
\cite{lee2021realtime} additionally add hydraulic limits in power usage and actuation force for machine adaptability.
More detailed soil-tool interaction models improve the tracking performance during the contact-rich sub-surface part of the trajectory as shown by \cite{sotiropoulos2021koopman}.
For accurate force modeling, soil parameters can be estimated from machine data for use in analytical earthmoving or \ac{FEE} models \cite{wagner2023soilestimation}, which can then support computationally efficient trajectory planning \cite{abdolmohammadi2025wheelloader}.
These methods can explicitly adapt to physical limits of the machine, making them suitable candidates for retrofitting.
Soil mechanics are usually not included in planning-based techniques, leaving interaction-heavy behavior beyond bucket filling to manually engineered motions.

\subsection{Learning-Based Excavation and Earthmoving}
\label{subsec:sota_learning_excavation}
Learning-based control enables the use of more complex soil models at training time as well as enables highly soil adaptive and reactive behaviors during inference.
In end-to-end approaches, machine-specific controllers map observations directly to low-level actuation commands.
With that architecture, \cite{chen2024exact} use imitation learning on recorded reach-dig-dump cycles.
Collecting data from simulation and training with \ac{RL}, more advanced tasks such as rock capture~\cite{molaei2025rocks} or the combination of multiple motions~\cite{zhai2025ext} including those from demonstrations can be achieved.
Other systems retain classical control elements and add learned elements for difficult contact, as for example in jamming-prone terrain~\cite{franceschini2024oscillatory}, or a learned policy above conventional low-level tracking for boulder excavation~\cite{gruetter2025boulder}.
Integrating a more detailed analytical resistive-force model~\cite{Egli22SoilAdaptiveExcavation} uses end-to-end \ac{RL} for soil-adaptive bucket filling.
A similar task on a full-sized wheel loader was trained inside a learned latent model of the machine by \cite{eriksson2024dreamloader}.
\cite{spinelli2025materialhandling} show grasp-point selection on a \SI{40}{\tonne} material handler learned using a bulk material simulator based on cellular automation in order to simulate the collapse and pile redistribution.
Longer horizon planning tasks also benefit from higher fidelity soil simulation, as shown by~\cite{miron2022grading}, who uses a height field simulation to learn surface flattening on a scaled dozer.

The main limitation in all methods is the available terrain state representation.
While specific heuristic simulations work for the designed use case (e.g., bucket filling or object extraction), they do not generalize to other motions in contact with deformable soil.
Tasks are solved by handcrafted motions, demonstrations, or guidance rather than being inferred from the simulator state.
Beyond the height field, little terrain state mechanics are present, and they are not exposed to the learning algorithm for contact-rich redistribution tasks.

\subsection{Particle and Granular Simulation for Robotic Soil Interaction}
\label{subsec:sota_particle_simulation}
The aforementioned limitations of surface evolution, history-dependent effects, and high-fidelity tool-soil contact seen in heuristic models can be overcome using particle-based simulators.
For our simulator, we use widely adopted numerical methods for soil mechanics simulation including \cite{sulsky1994mpm}, elastoplastic constitutive modeling \cite{stomakhin2013snow}, affine particle-grid transfers \cite{jiang2015apic,jiang2016apic,hu2018mlsmpm}, and grid-based Coulomb-friction contact, through multiple velocity fields \cite{bardenhagen2000granular} or a merged field with collision particles \cite{han2019frictional}.
Within the domain of robotics, particle-based simulators have been employed to learn deformable object manipulation as well as used within optimization and validation. 
Recently, they are integrated into commercial robotic simulators such as Newton~\cite{newton2026,daviet2026mixedmpm} or Algoryx~\cite{servin2021agxterrain}.

Learning deformable object manipulation on particle simulators has been employed for tabletop testbeds.
Differentiable simulation makes the material state itself the manipulation objective, as shown by~\cite{huang2021plasticinelab} for elastoplastic materials and for fluids and multi-material scenes by~\cite{xian2023fluidlab}.
GPU multiphysics platforms extend this to coupled rigid-deformable systems for large-scale policy training \cite{xing2025rewarped},
and dedicated contact formulations make \ac{MPM}-rigid coupling stable enough for interactive simulation \cite{yu2025convexmpm}.
For granular media, simulating discrete particles and collisions on GPU is fast enough for training \ac{RL} policies \cite{millard2023granular} and can be calibrated from depth observations to predict piles and pattern formation \cite{matl2020inferring}, while shaping of sand-like media into target structures has been learned from height-map sand models \cite{kreis2025shaping}.
These systems are deployed on centimeter-scale plasticine, fluids, or dry sand in tabletop work boxes.

Furthermore, the simulated material state can serve as an optimization objective for material parameters or excavation trajectories as shown by~\cite{yang2026ddbot} and validated on a tabletop arm.
Granular repositioning has been formulated as trajectory optimization over a learned surrogate of a \ac{MPM} solver \cite{aoyama2024granularcontrol},
and differentiable \ac{MPM} has been applied to deformable-object control more generally by \cite{bolliger2025diffmpm}.
Gradients for differentiable methods are acquired through automatic differentiation through a physics simulator or learned surrogate.
Where both paradigms were compared on one simulator, gradient-based control was stronger on short horizons and degraded on long-horizon, multi-stage tasks \cite{huang2021plasticinelab}.

For validation in virtual large-scale scenarios, particle-based simulators can be utilized for multiple elements of the simulation chain.
\ac{MPM} with nonlocal granular fluidity reproduces blade and wheel interaction forces in three dimensions \cite{haeri2022granularflow}, while
GPU \ac{SPH} terramechanics scales to terrain kilometers long with on the order of \(10^8\) particles~\cite{unjhawala2025crm}.
Differentiable \ac{MPM} in Warp \cite{macklin2022warp,daviet2026mixedmpm}, can be used for example for geomechanical inverse analysis \cite{zhao2026geowarp},
and 2D \ac{MPM} has been applied to bucket-soil interaction as an earthwork parameter study without plasticity or reported throughput \cite{kim2021soilbucket}.
These formulations report run times orders of magnitude slower than real time \cite{haeri2022granularflow,unjhawala2025crm}, and none couples a policy to the solver during training or exposes material state as a reward.
Real-time rates are reached by multiscale terrain models that resolve particles only in an active zone around the tool \cite{servin2021agxterrain}.
Such simulators have also been used to train policies using \ac{RL}~\cite{OzakiDozer2025}.
\cite{Aoshima_Servin_2024} show that moderate soil simulation fidelity can be sufficient, with a \SI{10}{\percent} sim-to-real gap in terms of trajectory, velocity, forces, load mass, and work.
A computationally efficient granular media model built for \ac{RL} likewise enabled quadrupedal locomotion on real sand \cite{choi2023deformableterrain}, and coarse-resolution simulation has been used to reduce the cost of reaching fine-resolution policy performance \cite{kadokawa2025progressive}.
\cite{kadokawa2026obstacle} use a low-fidelity particle simulator to learn obstacle removal for an \SI{11.5}{\tonne} excavator, with rewards defined on obstacle removal and bucket-to-obstacle distance.

This work aims to fill a literature gap in tool-soil interaction at machine scale, expose internal material states such as compaction as rewards, and provide throughput sufficient for model-free \ac{RL}.

\subsection{Transfer, Embodiment Dependence, and Field Deployment}
\label{subsec:sota_transfer}
Reliable localization, perception, force estimation, and low-level motion control are prerequisites for real-world deployments.
\cite{nubert2022graph} shows graph-based multi-sensor fusion for consistent localization of autonomous construction machines under field conditions.
End-effector force estimation from hydraulic pressures and inertial measurements allows a policy or controller to reason about the interaction loads actually applied at the bucket \cite{werner2025forcepayload}.

Transfer across embodiments remains an open problem due to significant differences in hydraulic actuation, dynamics, and geometry across commonly used machinery.
We instead take inspiration from manipulation research, where demonstrations recorded with handheld grippers are learned as end-effector-space policies and executed on robot arms through low-level controllers, decoupling the skill from the specific embodiment \cite{song2020graspingwild,chi2024umi}.
For platform independence, the learned policy operates in a normalized task space and we separate the machine- and hydraulics-specific parts into conventional controllers.

\section{Method and Deployment Architecture}
\label{sec:architecture}
\begin{figure*}
    \centering
    \includegraphics[width=\linewidth]{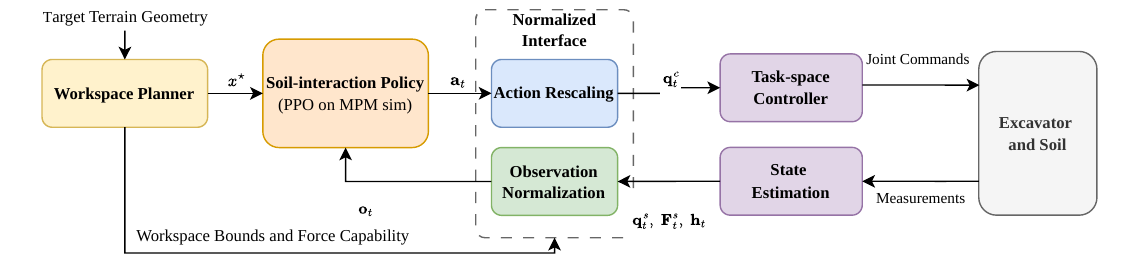}
    \caption{Overview of the working blocks in this paper in the context of task execution. The workspace planner selects local workspaces and the task target coordinate \(x^{\star}\), the soil-interaction policy learned in particle simulation acts in normalized task space, the normalized interface rescales observations and actions using the calibrated workspace bounds and force capability, and the low-level task-space controller tracks the commanded shovel pose. Green and blue mark observation and action processing, purple marks machine-specific nodes not detailed in this paper and orange the policy checkpoint.}
    \label{fig:systemOverview}
\end{figure*}

For deployment, we embed the local soil manipulation skill policy into a stack coordinated by a state machine.
Figure~\ref{fig:systemOverview} shows the hierarchy that executes a long-horizon task.
The workspace planner turns the requested terrain geometry into aligned local workspaces and attack points.
For the tasks presented, the planner acts purely feed-forward in coordinate space.
Inside each workspace, the learned policy performs the contact-rich in-soil motion without repositioning the excavator or actuating the cabin rotation joint.
A normalized interface converts between policy variables and physical quantities using the workspace bounds \(L_x, L_z\) and the force capability \(F_{\max}\).
The machine-calibrated low-level controller then tracks the resulting Cartesian shovel target \(\mathbf{q}^{c}_t\).
Our contribution is the learned soil-interaction policy, a particle-based simulator it is trained in, and the normalized interface that transfers the controller between machines.
Workspace planning, state estimation, and low-level tracking are realized with conventional non-learning-based components.
Since only the interface and low-level controller are machine-specific, the same policy checkpoint can be deployed on various embodiments without retraining, by adapting only the calibrated interface layer.

\subsection{Single-Workspace Task-Space Policy}
\begin{figure*}
    \centering
    \includegraphics[width=0.8\linewidth]{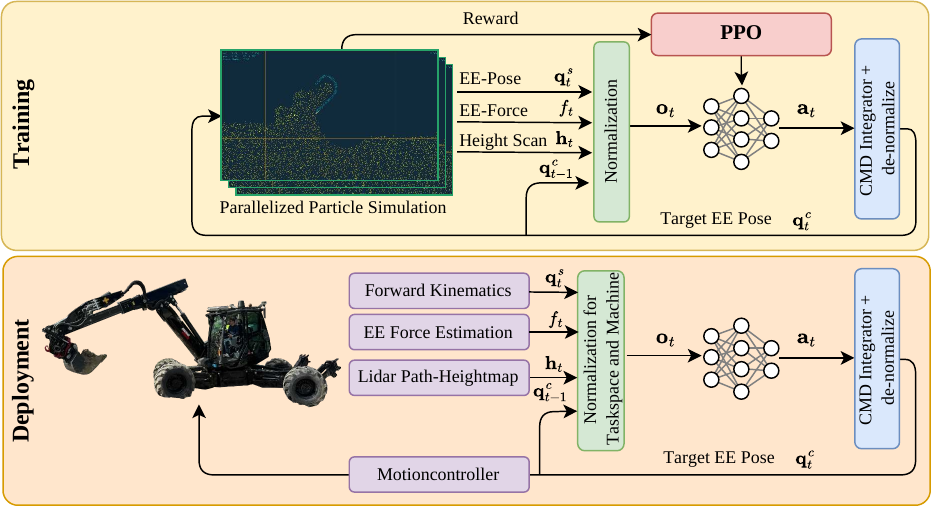}
    \caption{Training and deployment architecture for the normalized single-workspace soil-manipulation policy. Green and blue indicate observation and action pre- and post-processing steps adapted to workspace size and machine capability. Purple elements are machine-specific interface nodes. The same policy checkpoint is reused across platforms only after workspace, bucket, perception, force-normalization, and tracking interfaces are calibrated.}
    \label{fig:wholeSetup}
\end{figure*}

The learned controller acts inside one aligned vertical workspace and is formulated in Cartesian task space.
On heavy hydraulic machines, the cabin rotation joint is less suitable for contact-rich lateral soil manipulation than the in-plane boom, stick, and bucket joints.
The workspace planner therefore uses the cabin joint to align the workspace, while the policy commands 2D shovel motion in the arm plane with all high-torque actuators of the machine, matching the workflow of expert operators.
Figure~\ref{fig:wholeSetup} shows the training and deployment setup of the components from Figure~\ref{fig:systemOverview}.

\subsubsection{Actions}
\label{sub:actions}
At each control step, the policy action $\mathbf{a}_t$ is added to the Cartesian shovel target in the two-dimensional arm plane $(x, z)$,
represented by a control-point position and shovel angle of attack $\Theta$.
Updating a leading target pose before it is reached mitigates the jerky motion usually observed with end-effector space control.
\begin{equation}
\mathbf{a}_t =
\begin{bmatrix}
a^x_t & a^z_t & a^\Theta_t
\end{bmatrix}^{\top},
\qquad
\mathbf{q}^{c}_{t+1}=\mathbf{q}^{c}_{t}+\Delta t_c\,\mathbf{a}_t,
\label{eq:task_space_policy_update}
\end{equation}
where \(\mathbf{q}^{c}_t=(x^c_t,z^c_t,\theta^c_t)\) denotes the control point and shovel orientation given to the low-level controller, and \(\Delta t_c\) is the policy control period.
The low-level machine controller tracks this command as described in Section~\ref{sub:deployment}, effectively removing hydraulic joint control and inverse kinematics from the learning objective.
Its implementation decides the \ac{EE} target $\mathbf{q}^{c}_t$ tracking speed, allowing on-machine tuning for a desired and safe behavior.
Safety constraints on machine motion can be enforced deterministically in the motion control layer.

\subsubsection{Observations}
\label{sub:observations}
We keep the observation independent of the excavator embodiment and control dynamics.
All components are normalized, which improves training and eases transfer between machines.
Horizontal and vertical quantities are scaled by the local workspace lengths \(L_x\) and \(L_z\), the shovel orientation by \(2\pi\), and the shovel contact force \(\mathbf{F}^{s}_t\) by the machine force capability \(F_{\max}\), which also sets the maximum desired force in the reward function.
A tilde marks the normalized quantity,
\begin{subequations}\label{eq:obs_norm}
\begin{align}
\tilde{\mathbf{q}}^{s}_t &=
\bigl(x^s_t/L_x,\; z^s_t/L_z,\; \Theta^s_t/(2\pi)\bigr),
\label{eq:obs_norm_shovel_pose}\\
\tilde{\mathbf{q}}^{c}_{t-1} &=
\bigl(x^c_{t-1}/L_x,\; z^c_{t-1}/L_z,\; \Theta^c_{t-1}/(2\pi)\bigr),
\label{eq:obs_norm_target_pose}\\
\tilde{x}^{\star} &= x^{\star}/L_x,
\label{eq:obs_norm_target_coordinate}\\
\tilde{f}_t &= \|\mathbf{F}^{s}_t\|_2/{F}_{\max},
\label{eq:obs_norm_force}\\
\tilde{\mathbf{h}}_t &= \mathbf{h}_t/L_z,
\label{eq:obs_norm_height}
\end{align}
\end{subequations}
where \(\mathbf{q}^{s}_t=(x^s_t,z^s_t,\theta^s_t)\) is the measured shovel pose, \(x^{\star}\) the task target coordinate which is the input into the controller where to place the center of the embankment in the workspace, and \(\mathbf{h}_t\) the sampled terrain height profile.
The observation concatenates the 2D shovel pose \(\tilde{\mathbf{q}}^{s}_t\in\mathbb{R}^{3}\), the previous Cartesian target pose \(\tilde{\mathbf{q}}^{c}_{t-1}\in\mathbb{R}^{3}\), the task target coordinate \(\tilde{x}^{\star}\in\mathbb{R}\), the force utilization \(\tilde{f}_t\in\mathbb{R}\), and the terrain height profile \(\tilde{\mathbf{h}}_t\in\mathbb{R}^{N_h}\) with \(N_h=30\),
\begin{equation}
\mathbf{o}_t = \left[\tilde{\mathbf{q}}^{s}_t,\; \tilde{\mathbf{q}}^{c}_{t-1},\; \tilde{x}^{\star},\; \tilde{f}_t,\; \tilde{\mathbf{h}}_t\right]^{\top}\in\mathbb{R}^{38}.
\label{eq:observation_vector}
\end{equation}
Terrain height samples are updated only when the shovel is above a threshold height, mimicking the inability to perceive terrain occluded by the shovel.
The force utilization indicates whether the current interaction is close to the calibrated machine capability, allowing the policy to adapt to machines with different force limits and to soils of varying resistance.
Only its magnitude enters the observation, whereas the reward and termination conditions in Sections~\ref{sub:rewardFunctions} and~\ref{sub:termination} use the individual components.
The magnitude is sufficient to indicate proximity to stalling, and the direction of the interaction follows from the commanded motion relative to the observed terrain.
During training, we add noise on the force and pose observation for policy robustness.

\subsubsection{Training}
The policy is trained on the local workspace simulation, which contains the soil and free-floating shovel controllable through a simplified command-following interface.
Learning is thereby reduced to discovering how the shovel should move through soil to achieve a material objective without any knowledge about the embodiment embedded into the policy.
Instead of hand-crafted motions, our rewards are computed from simulated material state, which enables the policy to optimize the desired physical outcome.

\subsubsection{Deployment}
\label{sub:deployment}
During deployment, a policy interface layer defines the physical workspace bounds, transforms the measured terrain profile into the policy coordinate system, rescales the commanded Cartesian shovel pose, and computes the normalized force utilization from estimated end-effector forces and the current machine capability.
We use a \ac{MPC}-based end-effector motion controller from~\cite{Werner2026} to track the rescaled shovel targets on hydraulic machinery.
A stiff motion controller is required for this method.
If the target lies slightly inside resistant soil, the machine should apply the available force needed to approach the target, subject to safety limits.
A purely compliant controller that reduces force in proportion to target error would change the physical meaning of the policy action.
Calibrating these interfaces requires conventional methods only, while our learned checkpoint applies to machines with comparable bucket geometry and compatible workspace aspect ratios.

\subsection{Worksite Supervisor}
\label{sub:stateMachine}
Most earthmoving objectives require a sequence of local interactions with machine repositioning in between.
A high-level state machine decomposes the requested geometry into aligned workspaces of approximately one shovel width and drives the machine to the associated base poses.
For each attack point it moves the arm to an intermediate pose that leaves the workspace unobstructed for the terrain scan, places the shovel in contact, and runs the learned policy until its terminal condition, for example when the shovel exits above a prescribed height after completing an embankment stroke.
We use the same structure for all tasks in this paper and only change the task-specific completion criteria.

\section{Simulator and Policy Training}
The control policy is learned with batched on-policy model-free reinforcement learning in a custom \ac{MPM}-based simulator.

\subsection{Soil Simulation}
\label{sub:soilSimulation}
The soil is simulated in the vertical arm plane with a two-dimensional explicit \ac{MPM} solver, implemented using NVIDIA Warp~\cite{macklin2024warp}.
We batch over environments and parallelize the \ac{MPM} update on GPU.
General-purpose 3D solvers are impractical at the domain size, particle count, and sample rate required for this task, and they do not expose the material-state variables we use for rewards.
The soil is treated as an elastoplastic continuum with a Drucker-Prager return mapping in logarithmic strain space~\cite{klar2016drucker}.
We introduce two earthworks specific additions used in the rewards.
First, the shovel is a separate set of particles following a prescribed rigid motion and is coupled to the soil by a dual-grid contact scheme with Coulomb friction.
Accumulating the contact impulses yields the shovel interaction force \(\mathbf{F}^{s}_t\) that we can use in observation and reward formulations.
Second, every particle carries a persistent plastic volumetric compaction memory \(\nu_p\) that records mostly hydrostatic plastic compression, hardens the material, and provides the material-state signal for compaction related tasks.
We use a simulation domain of \SI{5.0}{\m}\(\times\)\SI{3.0}{\m} on an \(80\times48\) grid with a physics time step of \(\Delta t=2\times10^{-3}\,\mathrm{s}\) and a \SI{10}{\Hz} control rate, which reaches about 533 times real time with 1500 parallel environments with \(7000\) soil particles each on a single NVIDIA GH200.
\subsection{Solver Additions and Throughput}
\label{sub:solverDetails}
Beyond the compaction memory and the shovel coupling, two modifications of the standard Drucker-Prager update are needed for stable and realistic behavior at this resolution.
First, we cap the frictional contribution of the yield criterion,
\begin{equation}
f(\boldsymbol{\varepsilon})
=
\left\lVert \hat{\boldsymbol{\varepsilon}} \right\rVert
-
\frac{k_c}{2\mu}
-
\min\!\left(
-\frac{2\lambda+2\mu}{2\mu}\,
\alpha \operatorname{tr}(\boldsymbol{\varepsilon}),
\;
\beta\,\frac{k_c}{2\mu}
\right),
\label{eq:dp_yield}
\end{equation}
with Lam\'e parameters \(\lambda\) and \(\mu\) (shear modulus), elastic logarithmic strain \(\boldsymbol{\varepsilon}\) and its deviatoric part \(\hat{\boldsymbol{\varepsilon}}\), cohesive offset \(k_c\), friction coefficient \(\alpha\), and cap factor \(\beta=0.5\).
Because the Drucker-Prager cone is open in compression, the frictional gain is otherwise unbounded.
A confined particle accumulates elastic strain, and on decompression the yield surface drops faster than the strain relaxes, so the return mapping releases the stored energy in a single step and destabilizes the explicit update.
The cap acts as a compressive cap on the cone: at the confinement levels reached in these tasks the shear strength saturates at \((1+\beta)\) times its unconfined value, which bounds the stored deviatoric strain and leaves the strength pressure dependent only in the weakly confined material near the free surface.
Second, the compaction memory \(\nu_p\) feeds back into the material.
It hardens stiffness, cohesion, and friction angle by a factor \(1+k_h\nu_p\), where \(k_h\) is the compaction hardening coefficient.
It enters the elastic volume ratio \(J^{e}_p=\min(1,\det(\mathbf{F}_p)\exp(\nu_p))\), where \(\mathbf{F}_p\) is the deformation gradient stored for particle \(p\).
This relaxes compressive rebound while the clamp prevents compaction from creating artificial tensile suction.
The compaction memory also gradually shrinks the particle reference volume toward the locally observed packed volume, so that compacted soil occupies less volume in subsequent steps.

Contact is resolved on the grid between separate soil and shovel velocity fields~\cite{bardenhagen2000granular}, with the contact normal estimated directly from the rasterized shovel particle cloud, which avoids a signed-distance representation and keeps the kernel parallel over grid nodes.
The interaction force \(\mathbf{F}^{s}_t\) accumulates normal and Coulomb-friction impulses over all physics substeps of a control step.
It therefore also captures the tangential resistance that dominates scraping and cutting.
On the real machine the reaction force is carried through the hydraulics and chassis into the ground, acting as a momentum sink.
\(\mathbf{F}^{s}_t\) therefore measures the effort that the commanded shovel motion demands from the machine, which is subsequently used in observation and reward formulations.

\begin{figure}
    \centering
    \includegraphics[width=0.8\linewidth]{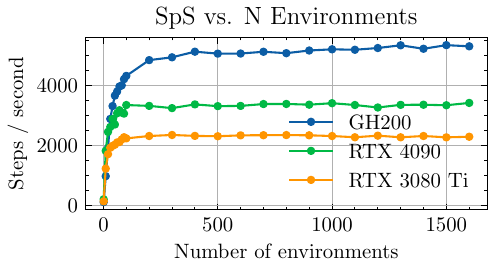}
    \caption{Control steps per second (\(\Delta t_c=0.1\,\mathrm{s}\)) versus number of environments. Each environment uses 7000 soil particles, 1000 shovel particles, grid spacing \SI{6.25}{\cm}, and physics time step \(\Delta t=2\times 10^{-3}\,\mathrm{s}\). Evaluation for a single NVIDIA GH200, RTX4090, and RTX3080Ti.}
    \label{fig:spsVnenvs}
\end{figure}

All particle and grid arrays carry the environment index in their leading dimension, and every kernel processes all environments simultaneously, parallelizing additionally either over particles, for the shovel velocity update and the particle-grid transfers, or over grid nodes, for grid clearing and the grid update with contact.
The compute therefore scales as \(\mathcal{O}(N_{\mathrm{env}}N_p)\) and \(\mathcal{O}(N_{\mathrm{env}}N_g)\) for \(N_p\) particles and \(N_g=N_xN_z\) grid nodes per environment.
We capture the full sequence of physics substeps between two control updates once as a Warp execution graph and replay it during training to remove launch overhead.
The particle-grid scattering transfers dominate the runtime on GH200.
Memory access accounts for about \SI{80}{\percent} of the runtime and the soil model for about \SI{20}{\percent}, while all remaining kernels are more than an order of magnitude faster.
Switching to gathering, custom data structures, and manual cache optimization did not improve this significantly, so we kept the straightforward scattering implementation.
Figure~\ref{fig:spsVnenvs} reports control steps per second over the number of environments for three GPUs.
With multi-GPU training on four GPUs we train at around 2000 times real time, and \ac{PPO} optimization time is negligible in comparison.

\subsection{Learning Soil Interaction}
\label{sub:learningSoilInteraction}
\begin{figure}
    \centering
    \includegraphics[width=0.8\linewidth]{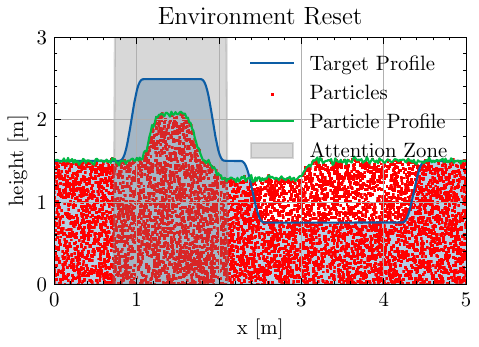}
    \caption{Example environment reset.}
    \label{fig:env_reset}
\end{figure}

The policy acts at the control rate \(\Delta t_c=0.1\,\mathrm{s}\), while the particle simulator advances at the physics time step \(\Delta t=2\times 10^{-3}\,\mathrm{s}\).
% Observations~\ref{sub:observations} and actions~\ref{sub:actions} are populated from the internal simulator state and the policy output.
The simulated shovel pose linearly follows an incrementally updated control pose, mimicking the end effector tracking on the real machines.
At reset, each environment draws a precomputed terrain, generated by randomly placing positive and negative features along the workspace and smoothing them into a physically viable profile.
Figure~\ref{fig:env_reset} shows an exemplary environment reset state.
We randomize the shovel spawn pose, the soil parameters across silt, sand, and clay and compaction states, the machine force capability \(F_{\max}\), the shovel speed limit, and observation noise.
By training across machines with different force limits \(F_{\max}\), the policy learns to adjust its stroke according to the capability specified by the deployment interface.

\subsection{Reward Functions}
\label{sub:rewardFunctions}
All rewards are computed at the control rate from the simulated material state.
Tasks share an action penalty \(r_{\mathrm{act}}=-\lambda_a\|\mathbf{a}_t\|_2^2\), a tracking penalty \(r_{\mathrm{track}}\) proportional to \(\|\mathbf{q}^{c}_t-\mathbf{q}^{s}_t\|_2\) once the commanded target drifts more than \(\delta_q\) from the simulated shovel pose, and a terminal reward \(r_{\mathrm{term}}\) from Section~\ref{sub:termination}.
We use a two-stage curriculum switched at a mean-return threshold.
The first stage adds a term \(r_{\mathrm{init}}\) keeping the commanded control point near a reference pose, the second removes it and opens the reset and observation randomization.

\emph{Embankment formation} and \emph{backfilling} have surface-shape objectives.
At reset we sample a target profile \(\mathbf{h}^{\star}\) and an attention mask \(\mathbf{m}\) and measure the attention-weighted profile distance
\begin{equation}
d_t =
\frac{\sum_{i}\mathbb{I}[m_i>m_{\min}]\,m_i\left|h_{t,i}-h^{\star}_{i}\right|}
{\sum_{i}\mathbb{I}[m_i>m_{\min}]\,m_i}.
\label{eq:profile_distance}
\end{equation}
Here, \(m_{\min}\) is the minimum attention weight above which a profile sample contributes to the distance.
The policy observes only the target coordinate \(x^{\star}\) and is unaware of \(\mathbf{h}^{\star}\) or \(\mathbf{m}\).
Therefore, its behavior is not conditioned on a soil-dependent cross-section collapse.
For embankment formation, \(\mathbf{h}^{\star}\) is a trench-and-embankment profile with a binary mask, and the reward is
\begin{equation}
r^{\mathrm{emb}}_t
=
\lambda_{d}\left[d^{\mathrm{best}}_{t-1}-d_t\right]_+
+ \lambda_{h}\!\!\sum_{i:\,h_i^{\star}>h_{\mathrm{emb}}}\!\!(h_{t,i}-h_{\mathrm{base}})
+ r_F + r_{\mathrm{sh}},
\label{eq:emb_reward_main}
\end{equation}
with \(d^{\mathrm{best}}_t=\min_{\tau\leq t}d_\tau\), \([x]_+=\max(x,0)\), \(h_{\mathrm{emb}}\) the target-height threshold selecting the raised embankment region, \(h_{\mathrm{base}}\) the reference height subtracted in the dense height bonus, and \(r_{\mathrm{sh}}\) collecting the shared terms.
Progress is rewarded only against the best distance seen so far, a small dense bonus acts on the raised part of the target, and \(r_F\) penalizes contact forces exceeding the sampled machine capability.
Rewarding a target profile localizes the reward to the intended region.
It proved more stable than rewarding added material or peak height and suppresses exploits based on transient unsupported height scans.
For backfilling, the target is a uniform level \(h_{\mathrm{flat}}\) with graded weights on the pile and the depression.
Progress is measured between consecutive steps with asymmetric weights, \(\lambda^{+}_d[\Delta_t]_+-\lambda^{-}_d[\Delta_t]_-\), \(\Delta_t=d_{t-1}-d_t\), where \([x]_-=\max(-x,0)\) is the magnitude of the negative part, so that moving material away from the target is penalized, and a Gaussian kernel around \(h^{\star}\) adds a dense bonus for samples already at level.
Since a flat profile can also be reached by leveling the pile locally, a transport term rewards the commanded horizontal movement \(x^{c}_t-x^{c}_{t-1}\) directed from the pile centroid \(\bar{x}^{+}_t\) to the hole centroid \(\bar{x}^{-}_t\), gated by the proximity of the shovel to the source and by the contact of the soil through the normalized force \(\|{\mathbf{F}}_t\|_2\).

\emph{Compaction} cannot be observed through surface shape.
The reward instead reads the mass-weighted mean plastic volumetric compaction memory \(\bar{\nu}_t\) of the particles in the attention zone \(\mathcal{A}\), and rewards improvement over the best value together with a dense term:
\begin{align}
r^{\mathrm{comp}}_t
&=
\lambda_{p}\left[\bar{\nu}_t-\bar{\nu}^{\mathrm{best}}_{t-1}\right]_+
+\lambda_{\nu}\bar{\nu}_t
-\lambda_{F}\left[\|{\mathbf{F}}_t\|_2-1\right]_+^2
\nonumber\\
&\quad
+\lambda_{N}\min(0,N^{\mathcal{A}}_t-N^{\mathcal{A}}_{t-1})
+ r_{\theta} - \lambda_{t} + r_{\mathrm{sh}},
\label{eq:comp_reward_main}
\end{align}
where \(\bar{\nu}^{\mathrm{best}}_t\) is the highest mass-weighted mean compaction reached so far in the episode after initial settling, \(N^{\mathcal{A}}_t\) is the particle count in the attention zone, so material pushed out of the zone is penalized, \(r_{\theta}\) penalizes commanded shovel angles outside the useful compaction range, and \(\lambda_t\) is a constant time penalty against stalling.

\subsection{Termination and Training}
\label{sub:termination}
All tasks share a timeout and a workspace bound check. 
Embankment formation additionally terminates successfully when the shovel exits the upper workspace region with a reward proportional to the accumulated profile improvement.
Backfilling positively terminates when the attention weighted profile distance falls below a tolerance. 
The remaining conditions are force based terminations that keep the learned strokes inside the sampled machine capability. 
For compaction, terminations prevent the policy from satisfying the objective by pulling or cutting into the material instead of applying pressure. This task is otherwise particularly prone to exploiting simulator dynamics.

Policies are trained with PPO using RSL-RL~\cite{rslrl} on 4000 parallel environments, with actor and critic multilayer perceptrons of three hidden layers of width 256.
Training a task takes on the order of \SI{3}{\hour} on four GH200 GPUs, with first signs of learning appearing after 50 learning iterations across all three tasks.
The common PPO parameters are discount factor \(\gamma=0.99\), GAE parameter \(\lambda=0.95\), PPO clip parameter \(0.2\), value-loss coefficient \(0.5\), clipped value loss, two learning epochs, eight mini-batches, fixed learning rate, and gradient-norm clipping at \(0.5\).
The setup uses about 5 GB memory per GPU.

\section{Evaluation}
\label{sec:evaluation}
For evaluation, we deploy our control stack from \Cref{sec:architecture} on the \SI{11.5}{\tonne} hydraulic M445 as well as the \SI{500}{\g} LeExcavator.
We quantitatively evaluate the embankment creation policy in a long horizon field trial and qualitatively show material state conditioning and adaptivity on compaction and backfilling experiments.
As we are not aware of an existing method that provides a fair baseline with the ability to use out-of-shovel material and change the soil compaction, we use a human expert for performance comparison.

\subsection{Embankment}
\label{sub:embankmentExperiment}
In the embankment experiment, we demonstrate the core contributions of the paper by leveraging out-of-bucket bulk displacement and material flow.
We evaluate it with a large-scale outdoor deployment and in a direct expert comparison on M445.
Menzi Muck M445 is an \SI{11.5}{\tonne} spider excavator with retrofitted kinematic sensing, LiDAR and cylinder pressure sensors.
End-effector forces are estimated from the hydraulic pressures~\cite{werner2025forcepayload} and the terrain profile from LiDAR points averaged along the shovel path.
Maximum force capability $F_\text{max}$ is determined by manually pushing into the ground with a flat shovel.
This value can be modulated by the current shovel pose to take the changes in linkage advantage across the workspace into account.

The \SI{42}{\m} target embankment path is discretized into overlapping scoops by a supervising planner.
It groups attack points into workspaces at base poses chosen from the reachable arm range. 
For statistical evaluation we use 3 policy executions per attack point for a higher embankment and to recover collapsed material from the first stroke (201 executions across 67 attack points).
Driving, chassis balancing over uneven terrain, and path planning use conventional controllers.

\subsubsection{Field Deployment}
\begin{figure}
    \centering
    \includegraphics[width=\linewidth]{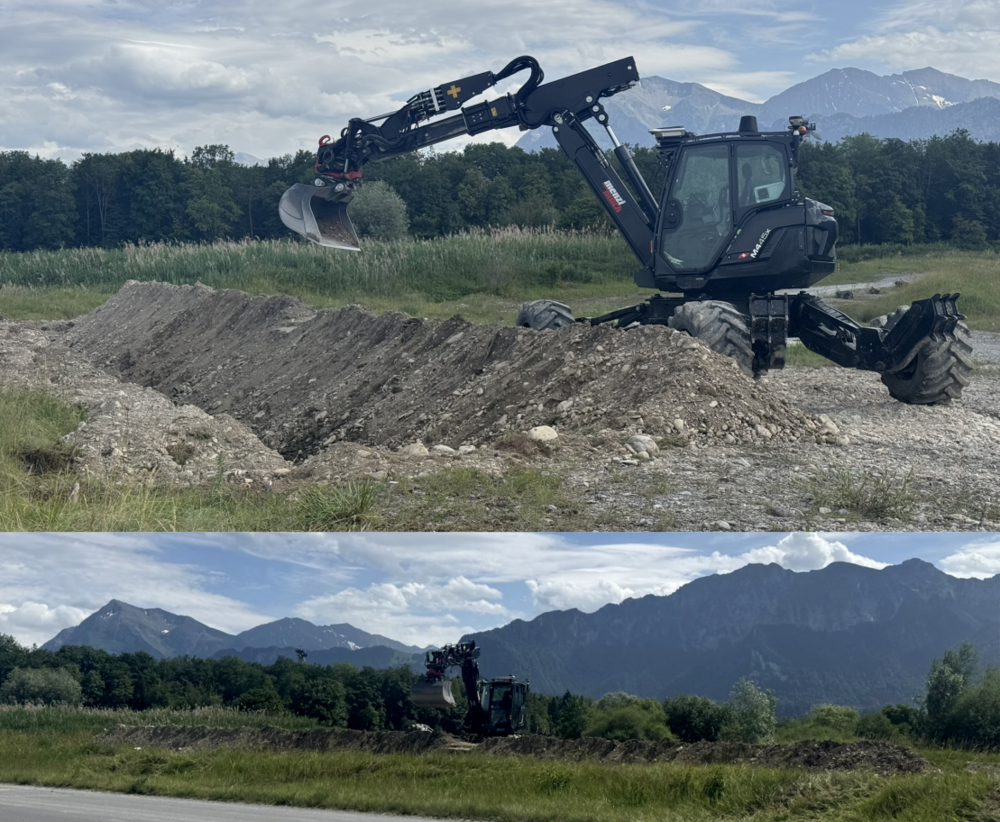}
    \caption{Autonomous construction of a \SI{42}{\m} embankment by M445 in one continuous mission of about \SI{45}{\min}. The measured valley-to-peak height is \SI{2.1}{\m}, and the advance speed is about \SI{1}{\m\per\min}.}
    \label{fig:bigWall}
\end{figure}

The autonomous system creates a \SI{42}{\m} long embankment at about \SI{1}{\m\per\min} (\SI{45}{\min} of continuous operation), encountering soil of different compaction and embedded rocks up to \SI{40}{\cm} in diameter.
The resulting structure, shown in Figure~\ref{fig:bigWall}, is continuous over the full length with a measured valley-to-peak height of \SI{2.1}{\m} despite non-homogeneous base material.
No human intervention or guidance beyond the globally registered coordinates of the target embankment were needed.
The reported mission was the first attempt at this scale. 
It is the only large-scale embankment run executed, with no discarded prior attempts.

\subsubsection{Per-Stroke Statistics}
\label{subsub:perStrokeStats}
The mission decomposes into 201 policy executions across 67 attack points grouped into 14 workspaces, with autonomous driving repositioning the machine.
Each rollout exits at the intended termination with no in-soil stalls, short-contacts or human intervention.
Every policy execution lasts $9.6 \pm 0.7$\,\si{\s} (range \SIrange{8}{11.9}{\s}) and reaches a peak shovel interaction force of $59 \pm 9.5$\,\si{\kN} (range \SIrange{36}{75}{\kN}).
We compute the displaced volume from a before/after 3D LiDAR scan.
Over the full mission the policy displaces \SI{110}{\m\cubed} of soil, with an average of \SI{0.55}{\m\cubed} per execution.

\begin{figure}
    \centering
    \includegraphics[width=0.8\linewidth]{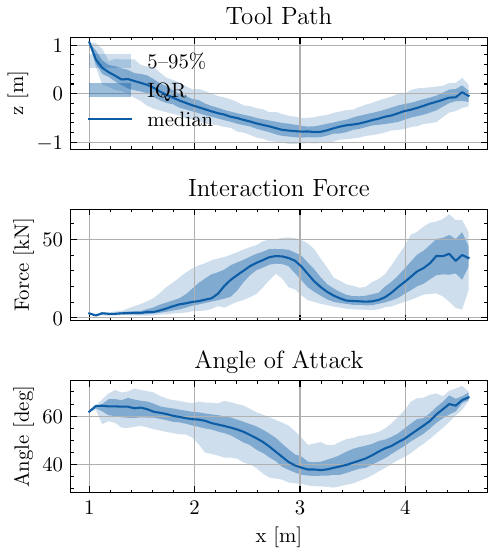}
    \caption{Shovel tool path, interaction force, and angle of attack as a function of stroke progress, aggregated over all 201 policy executions of the \SI{45}{\min} embankment mission. Shaded bands show the interquartile and \SIrange{5}{95}{\percent} range around the median.}
    \label{fig:fortification_stroke_envelope}
\end{figure}

Figure~\ref{fig:fortification_stroke_envelope} summarizes the shovel trajectory, interaction force profile, and angle of attack over all rollouts, reported as the median together with the interquartile and \SIrange{5}{95}{\percent} range as a function of shovel x position in the control workspace.
The envelope shows a consistent stroke shape across the mission.
The shovel penetrates at a high angle of attack, flattens while filling, and re-opens toward the exit.
The interaction force rises to a first local peak during penetration, eases during the fill phase, and rises again while pulling material up and out of the trench.
The interquartile band stays close to the median throughout the stroke, indicating a repeatable operation.
Outliers were observed when the policy had to adapt to buried rocks or hard soil.

\subsubsection{Expert Operator}
\label{subsub:expertOperator}
\begin{figure}
    \centering
    \includegraphics[width=0.8\linewidth]{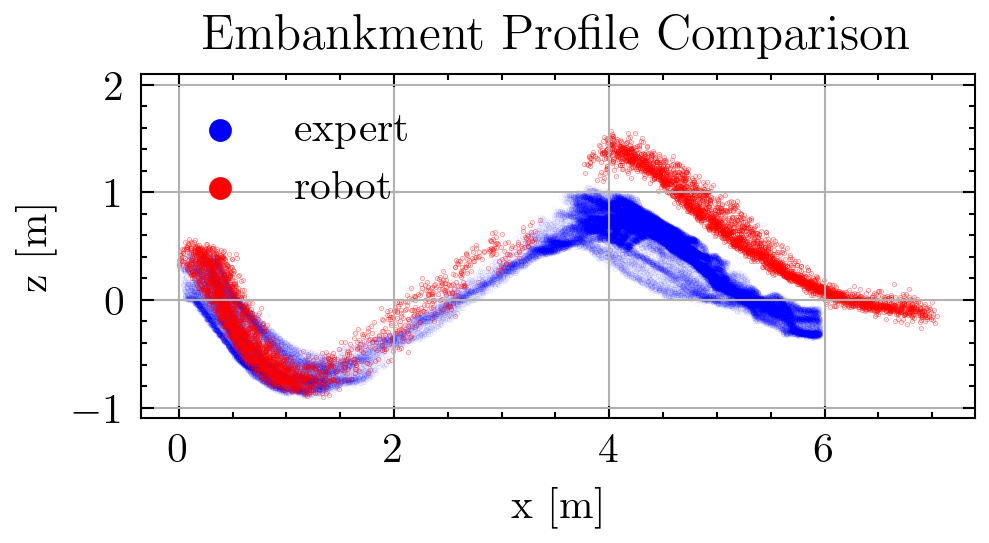}
    \caption{Same-setup \SI{9}{\m} embankment profile comparison between expert operator and autonomous controller. Expert-operator embankment height: \SI{1.46}{\m}. Autonomous-controller embankment height: \SI{2.11}{\m}. We project the point cloud along the embankment axis. Missing points on the robot data originates from point of view self-shadowing of the embankment.}
    \label{fig:profile_comparison}
\end{figure}

For a baseline comparison, we ask an expert spider-excavator operator with over 15 years of experience to create the same type of embankment next to our autonomously created one.
The expert is instructed to build the embankment as high as possible as quickly as possible, matching the objective for the autonomous controller.
We report the results as a single-trial comparison at one site and soil condition.
The expert chooses a similar approach to the autonomous controller, also using the excess material in front of the bucket by following a similar piling motion.
The main difference is the angle of attack with the expert using more bucket motion and lower angles of attack.
The expert creates an embankment of \SI{9}{\m} length in \SI{522}{\s}, resulting in a progression speed of \SI{1.03}{\m \per \min}, close to that of the autonomous operation. 
Figure~\ref{fig:profile_comparison} shows a comparison of the embankment profiles produced by the expert and the controller.
Both profiles are the projected point cloud of a \SI{9}{\m} embankment segment along the main axis. 
Variations in point density along the path originate from the oblique viewing angle of the scanner.

We measure valley to peak height by averaging the height along the x-direction and then comparing the highest average with the lowest.
The expert-operator embankment has a height of \SI{1.46}{\m}, while the autonomous-controller embankment reaches \SI{2.11}{\m}.
The autonomous controller also achieves lower height variation across the \SI{9}{\m} patch with a standard deviation of \SI{9}{\cm}, versus \SI{11}{\cm} from the expert operator.

\subsubsection{Conclusion}
With the large scale embankment experiment, we validate the effectiveness of our policy learning pipeline leveraging out-of-bucket material displacement at full scale.
Across 201 rollouts, the M445 built a \SI{42}{\m}, \SI{2.1}{\m} high embankment in \SI{45}{\min} without intervention, with tight per-stroke envelopes despite heterogeneous soil and buried rocks.
The controller matches expert operator progression speed, building a higher embankment at lower height variation in our trial.

\subsection{Validation}
In this subsection, we address the claims of this paper through qualitative experiments on hardware and ablation studies in simulation.

\vq{sub:platforms}{Can we adapt across excavators?}
\label{sub:embankment_leExcavator}
To evaluate calibrated cross-platform transfer, we run the same embankment policy from~\Cref{sub:embankmentExperiment} on the tabletop LeExcavator without policy retraining.
LeExcavator is based on the LeRobot SO-101 by~\cite{knight2024standard_open_so100_so101} with a custom shovel~\ac{EE}, uses an Azure Kinect DK depth camera for the height scan~\cite{microsoft2021azure_kinect_dk_hardware}, and computes \ac{EE} pose and force from servo position and current.

\Cref{tab:platform_calibration} lists the quantities calibrated per platform for the interface layer from~\Cref{sec:architecture}. 

\begin{figure}[t]
    \centering
    \subfloat[Menzi Muck M445\label{fig:M445}]{%
        \includegraphics[width=0.48\linewidth]{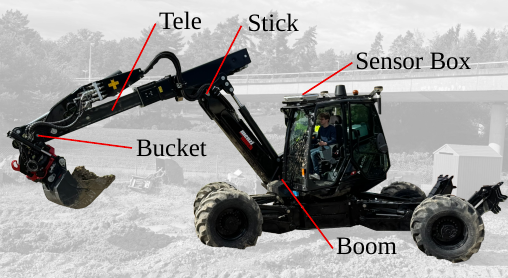}}
    \hfill
    \subfloat[LeExcavator\label{fig:LeExcavator}]{%
        \includegraphics[width=0.48\linewidth]{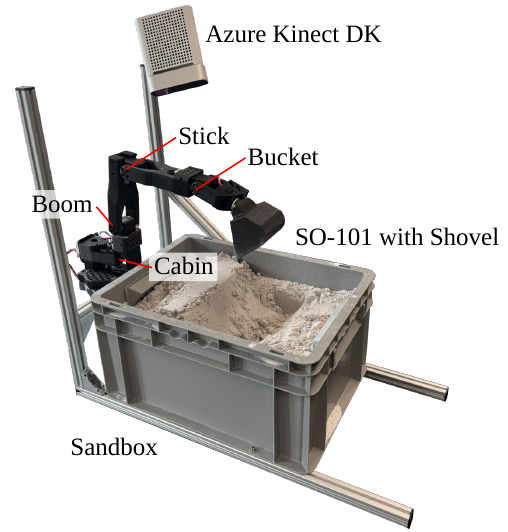}}
    \caption{The two evaluation platforms with link and joint naming. (a) Menzi Muck M445 robotic excavator: the arm links include joint position and velocity sensing, the hydraulic cylinders carry A- and B-side pressure sensors, and a LiDAR provides spatial perception. (b) LeExcavator, a small tabletop platform with electric servo motors for soil manipulation testing. The software interface matches that of the M445.}
    \label{fig:platforms}
\end{figure}

\begin{figure}
    \centering
    \includegraphics[width=0.8\linewidth]{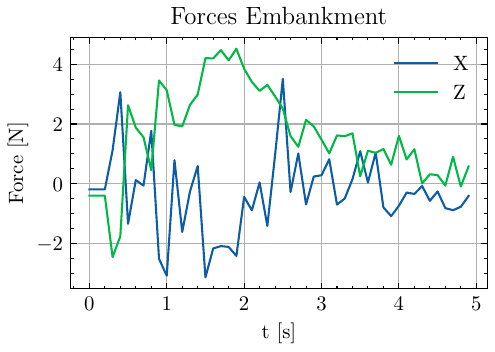}
    \caption{End-effector force estimates on LeExcavator with the embankment policy. The single-pass trace is noisier than the M445 hydraulic estimate but shows the qualitative contact sequence.}
    \label{fig:fortify_leexcavator_forces}
\end{figure}

\Cref{fig:platforms} shows both platforms with the link and joint naming.
The \SI{11.5}{\tonne} M445 is controlled by an on-board computer through the manufacturer's remote-control interface.
RTK GNSS, IMUs and LiDAR is used for machine pose estimation using~\cite{nubert2022graph}.
EE pose is provided by a commercial kinematic sensing system, computed from IMU measurements on the arm links.
A LiDAR in the sensor box provides terrain profile measurements.
EE interaction forces are estimated from cylinder pressure sensors and inertial readings using \cite{werner2025forcepayload}.
In LeExcavator, an Azure Kinect DK is chosen for height scanning due to the high resolution and low noise~\cite{10632150}.
EE pose and force are computed through forward kinematics and dynamics from servo position and current. 
Although servo current measurements are very noisy, experiments still succeeded, but cannot be evaluated quantitatively, as shown in Figure~\ref{fig:fortify_leexcavator_forces}.
\begin{table*}[t]
    \centering
    \scriptsize
    \caption{Per-platform calibration of the normalized interface.}
    \label{tab:platform_calibration}
    \renewcommand{\arraystretch}{1.15}
    \begin{tabular}{@{}p{0.30\linewidth}p{0.32\linewidth}p{0.28\linewidth}@{}}
        \toprule
        Calibrated quantity & Menzi Muck M445 & LeExcavator \\
        \midrule
        Machine mass & \SI{11.5}{\tonne} & \SI{500}{\gram} \\
        Workspace \(L_x \times L_z\) & \SI{5.5}{\m} \(\times\) \SI{3.3}{\m} & \SI{0.30}{\m} \(\times\) \SI{0.18}{\m} \\
        Workspace placement & \SI{2.5}{\m} to \SI{8.0}{\m} from the cabin & \SI{0.10}{\m} to \SI{0.40}{\m} from the base \\
        Force capability \(F_{\max}\) & \SI{100}{\kN} nominal & \SI{40}{\N} \\
        Force estimation & hydraulic cylinder pressures and \ac{IMU}~\cite{werner2025forcepayload} & servo position and current with forward dynamics\\
        \bottomrule
    \end{tabular}
\end{table*}

\begin{figure}
    \centering
    \includegraphics[width=\linewidth]{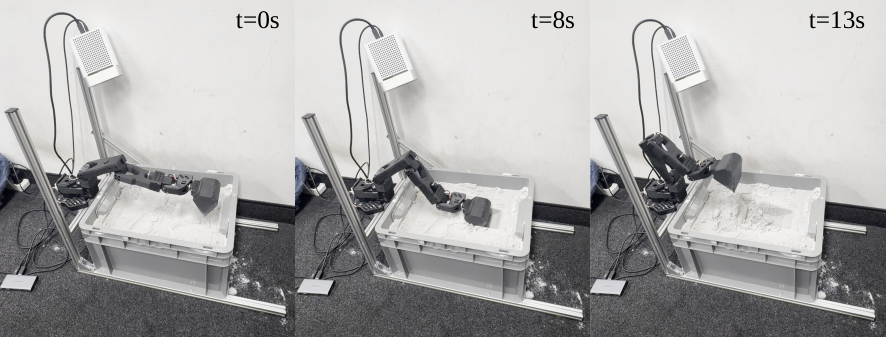}
    \caption{Single stroke of the embankment-creation policy on LeExcavator. The same checkpoint and weights are used without policy retraining. Only measured interface calibration changes.}
    \label{fig:fortify_leexcavator_mosaic}
\end{figure}

\begin{figure}
    \centering
    \includegraphics[width=0.8\linewidth]{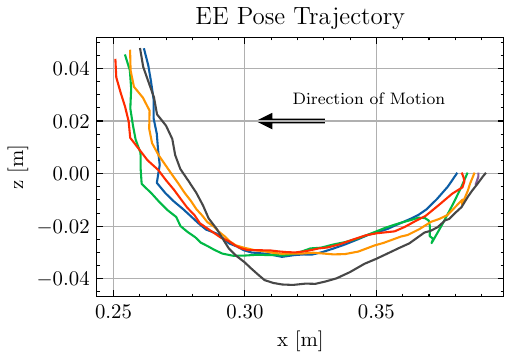}
    \caption{LeExcavator shovel trajectories with the embankment policy using the same checkpoint and measured interface calibration.}
    \label{fig:fortify_leexcavator_trajectories}
\end{figure}

Figure~\ref{fig:fortify_leexcavator_mosaic} shows a single-stroke execution of the embankment-creation policy.
The shovel trajectories in Figure~\ref{fig:fortify_leexcavator_trajectories} show the same qualitative trajectory shape as observed on the large hydraulic excavator.
The policy succeeds creating an embankment with the performed motion.

In subsequent experiments, we also compare the task execution between M445 and LeExcavator on the backfilling and compaction tasks.
Both transfers confirm that the policy performs well on different embodiments when deployed through the calibrated interface.

\begin{figure*}
    \centering
    \includegraphics[width=1.0\linewidth]{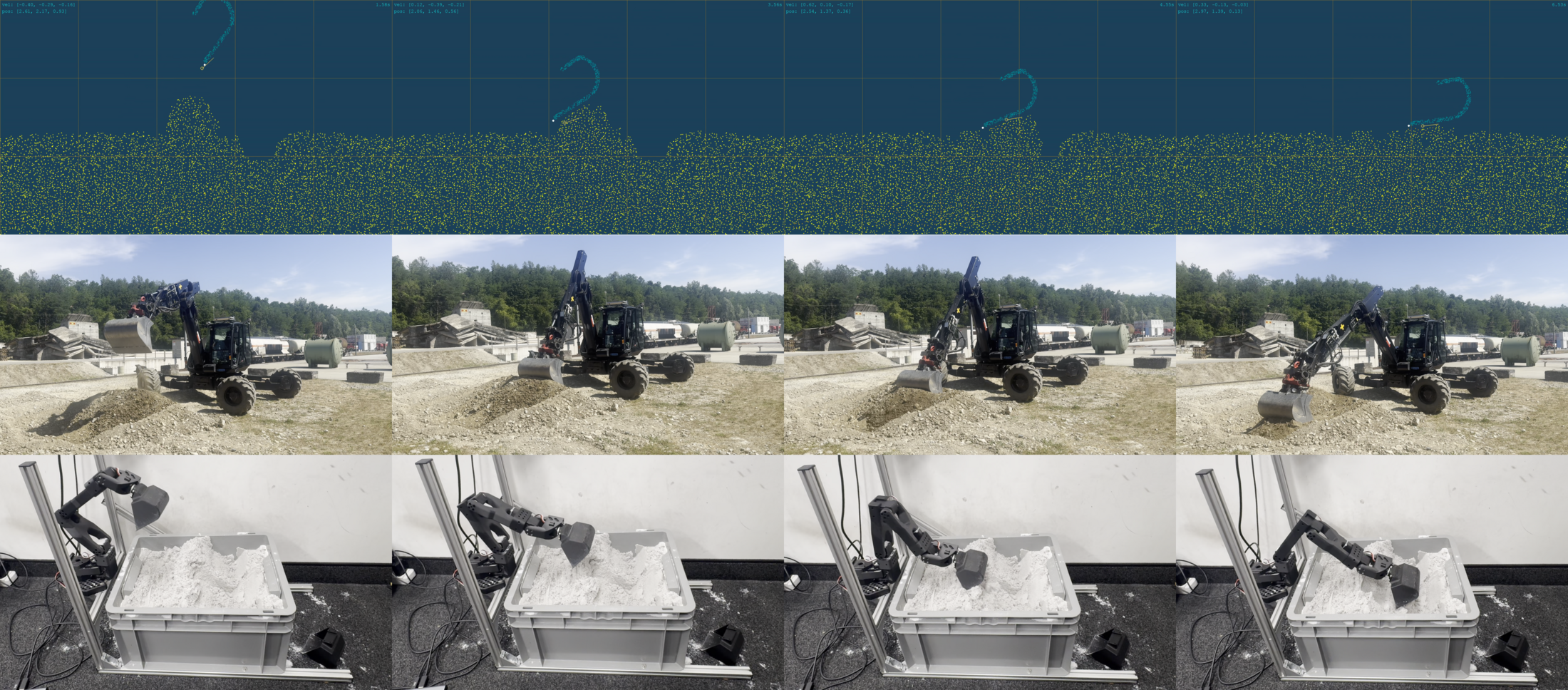}
    \caption{Representative backfilling shovel motion in the training environment, on M445, and on the tabletop LeExcavator. The policy uses the underside of the shovel by pushing material back into the trench at high angles of attack. The results are shown as qualitative transfer demonstrations.}
    \label{fig:backfill_mosaic}
\end{figure*}

\vq{par:backfilling}{Can we learn to use multiple faces of the shovel?}
The backfilling task reverses the effect of the embankment policy by pushing excess material into a pit.
Figure~\ref{fig:backfill_mosaic} shows representative shovel motions in the training environment, on M445, and on LeExcavator.
The learned policy uses multiple contact faces of the shovel, including the underside.
It moves the shovel at a high angle of attack in front of the embankment, pushes forward to displace material, and lowers the angle of attack near the end of the stroke to close the hole.
We qualitatively show that the controller is effective in backfilling a pit and that it transfers well between M445 and LeExcavator.

\begin{figure*}
    \centering
    \includegraphics[width=1.0\linewidth]{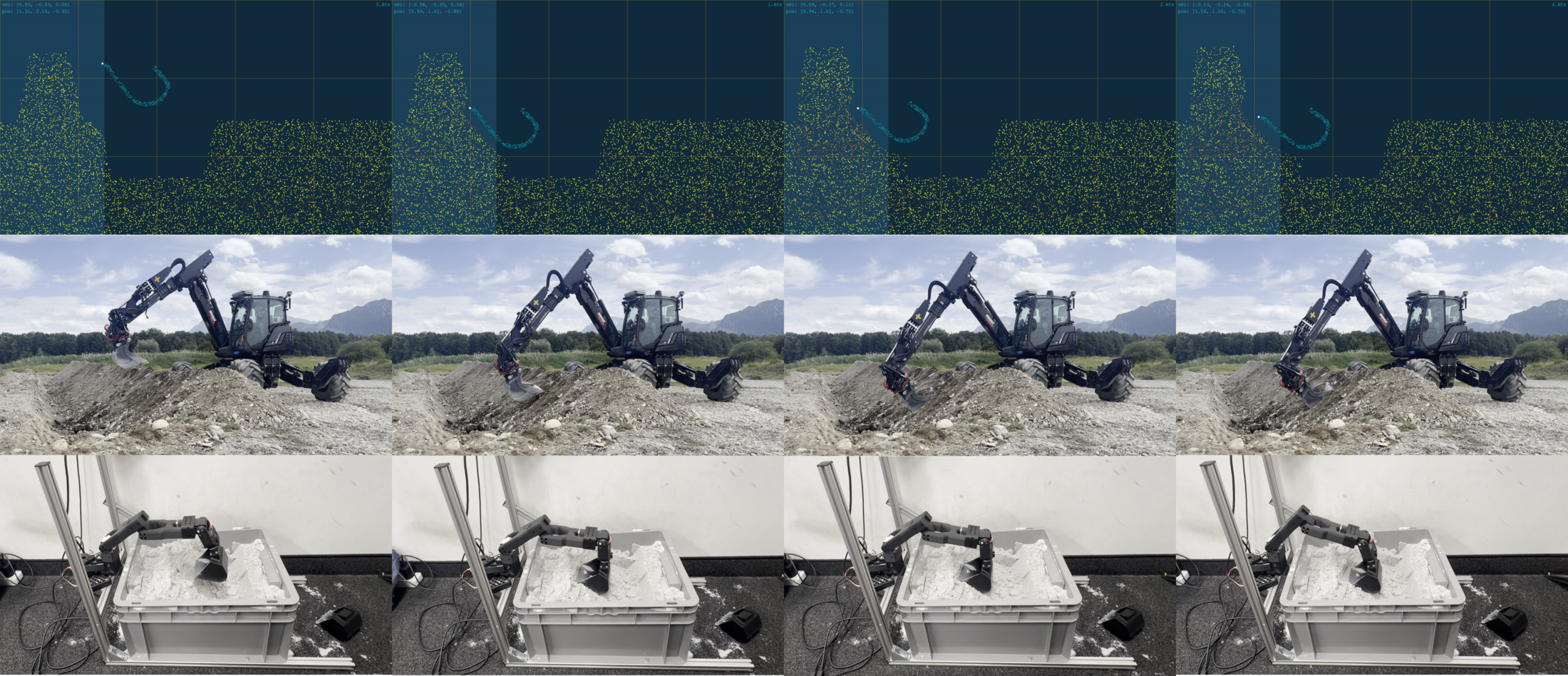}
    \caption{Representative compaction shovel motion in the training environment, on M445, and on LeExcavator. The policy approaches the embankment and applies force with the flat underside without visible collapse. Particle color indicates compaction state from yellow (loose) to red (compacted).}
    \label{fig:compaction_mosaic}
\end{figure*}

\begin{figure}
    \centering
    \includegraphics[width=0.8\linewidth]{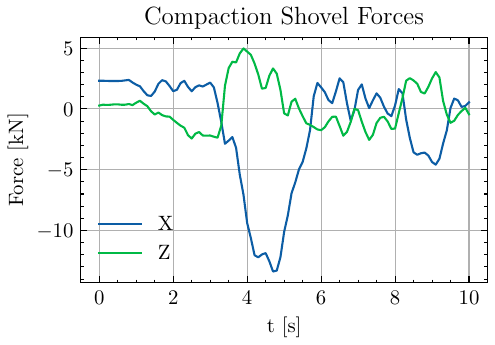}
    \caption{Shovel forces during the compaction skill on M445 for one representative run. The policy mainly exerts force in the x-direction while also pushing the material down.}
    \label{fig:compactionForces}
\end{figure}

\vq{par:compaction}{Can we condition the policy on the compaction of the soil?}
The compaction experiment qualitatively tests whether a reward on internal simulated particle state can yield a different contact behavior.
As described in Section~\ref{sub:rewardFunctions}, the reward encourages accumulation of compaction memory in the region of interest while penalizing particle displacement.
The learned policy uses the underside of the shovel to push against the embankment and applies force without collapsing the structure.
Figure~\ref{fig:compaction_mosaic} shows the learned motion in the simulation environment and on both real platforms.
Figure~\ref{fig:compactionForces} shows the interaction forces on M445.
On both platforms, the soil gets compressed without collapsing the pile.

\begin{figure}
    \centering
    \includegraphics[width=0.8\linewidth]{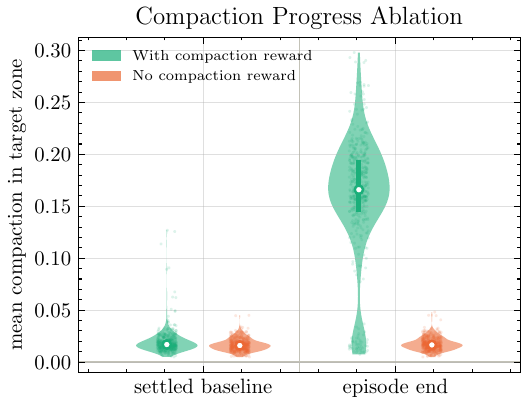}
    \caption{Mean attention-zone compaction at the settled baseline and at episode end, over 500 simulated rollouts of a policy trained with and without the compaction terms of \eqref{eq:comp_reward_main}. Both policies start from the same baseline. Only the reward-trained policy (green) compacts further by episode end, while the ablated policy (orange) stays at baseline.}
    \label{fig:compactification_reward_ablation}
\end{figure}

\vq{par:compactionAbl}{Does the internal particle state reward enable compaction?}
To isolate the effect of the reward on compaction state from the shared action and force penalties, we train an ablated policy without the compaction terms of \eqref{eq:comp_reward_main} and compare 500 simulated rollouts of each checkpoint. 
We read the mean attention-zone compaction ($\bar{\nu}_t$) at a settled baseline.
Figure~\ref{fig:compactification_reward_ablation} shows that both policies start from indistinguishable setups but only the compaction-trained policy improves compaction substantially (\num{0.158} vs.\ \num{0.017}) with per-rollout gain distributions clearly separated.
Without a reward on compaction state, the policy never learns the compacting motion in this learning setup.

\begin{figure}
    \centering
    \includegraphics[width=0.8\linewidth]{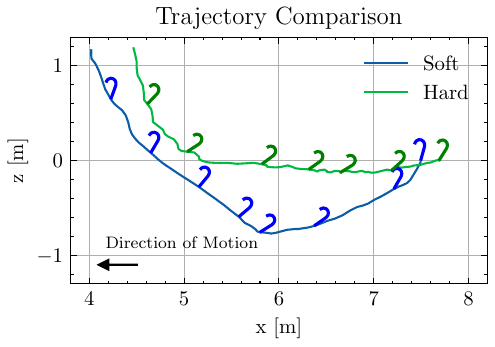}
    \caption{Comparison of shovel trajectories in hard and soft soil. Hard soil can only be scraped, while soft soil can be excavated. The realized trajectory changes visibly between the two material conditions.}
    \label{fig:trajectory_comparison_soft_hard}
\end{figure}

\begin{figure}
    \centering
    \includegraphics[width=0.8\linewidth]{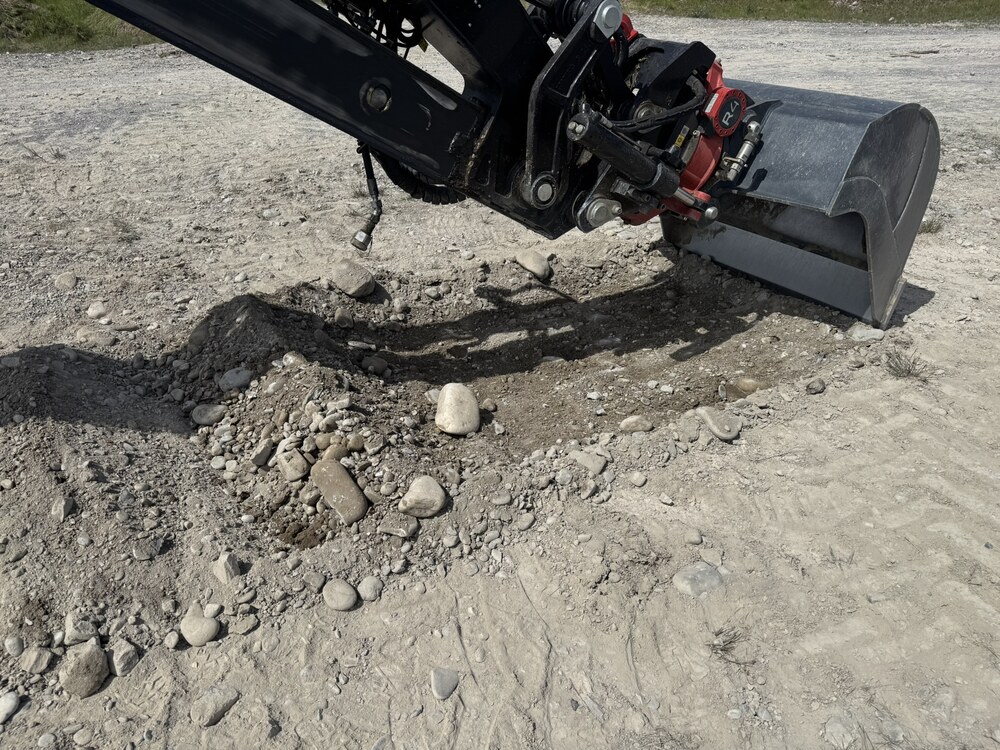}
    \caption{Hard soil after the scraping motion. Very little penetration and displaced material.}
    \label{fig:hard_scraping_soil}
\end{figure}

\vq{par:adaptSoilCond}{Can the policy adapt to different soil types?}
We test whether the learned policy changes its trajectory and force behavior based on the soil hardness.
The embankment policy is deployed on two flat workspaces, one with soft, excavatable ground and one on hard soil that requires scraping with high vertical force. 
Figure~\ref{fig:trajectory_comparison_soft_hard} compares the shovel trajectories of the soft and hard rollouts.
The soft-soil condition are later pictured in~\Cref{par:sim2realExperiments}, and Figure~\ref{fig:hard_scraping_soil} shows the hard soil after repeated scraping.
In soft soil, the controller penetrates at a higher angle of attack and reaches greater depth.
At the lowest point, the shovel partially closes to move material into the bucket and keep the shovel perpendicular to the direction of motion, maximizing the amount of displaced material in front of the bucket. 
This moves additional material that does not fit into the bucket up the embankment. 
During the pull-up, the shovel opens completely to dump the material onto the resulting embankment. 
On hard soil conditions, the shovel stays at a high angle of attack for scraping throughout the motion.
The policy achieves only shallow penetration even though it applies large vertical force to the ground.
In both cases, the machine does not get stuck and executes the full motion without human intervention.
The policy adapts the strategy effectively based on the soil condition.

\begin{figure}
    \centering
    \includegraphics[width=0.8\linewidth]{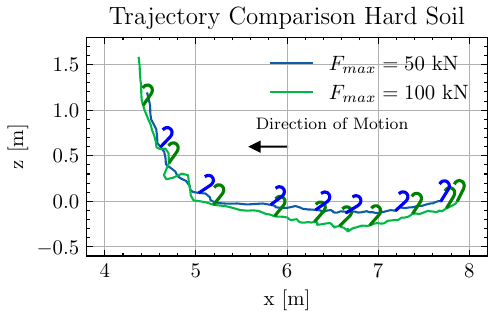}
    \caption{Comparison of shovel trajectories in hard soil with raw force-normalization values of \SI{50}{\kN} and \SI{100}{\kN}. The higher value leads to more applied force and about \SI{15}{\cm} deeper penetration in this representative test.}
    \label{fig:trajectory_comparison_capability}
\end{figure}

\vq{par:machineCapability}{Can the policy be transferred between machines of different force capability?}
The needed interaction force is highly dependent on the in-soil trajectory, which makes the policy responsible for adapting to the machine's capability.
In this experiment, we modulate value $F_{\max}$ of~\eqref{eq:obs_norm_force} and check the adapted behavior on very hard soil, which requires a scraping motion for displacement.
We test with $F_{\max}=\SI{50}{\kN}$ and \SI{100}{\kN}.
Figure~\ref{fig:trajectory_comparison_capability} shows the differences in the executed trajectory, with about \SI{15}{\cm} deeper penetration from the stronger normalization factor.

\begin{figure}
    \centering
    \includegraphics[width=0.8\linewidth]{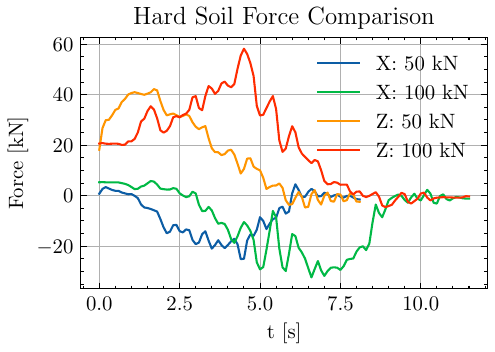}
    \caption{Shovel interaction force of the embankment policy during operation. The raw force-normalization values are \SI{50}{\kN} and \SI{100}{\kN}. The resulting force traces show the policy response to the force-utilization observation normalized at different values.}
    \label{fig:hard_soil_force_factor}
\end{figure}

The difference in behavior is visible in the \ac{EE} force profile in Figure~\ref{fig:hard_soil_force_factor}, with the stronger force-normalization setting pushing with about \SI{60}{\kN} into the ground and the weaker setting with about \SI{40}{\kN}.
The \SI{60}{\kN} rollout slightly lifts the front wheels of the machine, naturally limiting the maximum applied force.
The difference in applied force is also visible in the x-direction, with the stronger rollout applying more ripping force.

\begin{figure}
    \centering
    \includegraphics[width=0.8\linewidth]{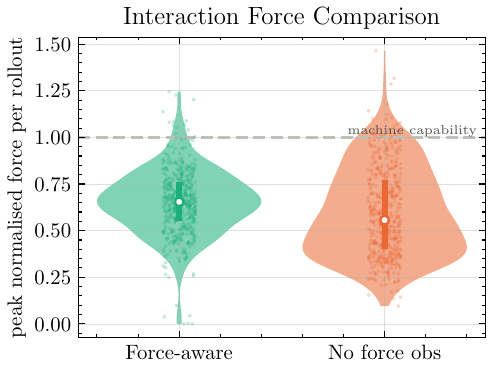}
    \caption{Peak interaction force per rollout, normalized by the machine-capability limit $F_{\max}$, over 500 simulated rollouts each for a policy with and without the force-utilization observation. Both checkpoints are evaluated at the same fixed $F_{\max}$, the highest capability value either policy has seen during training. The force-aware policy concentrates its force usage near the capability limit. The policy without the observation is more conservative on average and more widely spread, including rollouts that exceed the limit.}
    \label{fig:fortification_force_ablation_force_usage}
\end{figure}

\vq{par:forceObs}{Is end effector force observation necessary?}
The previous results show that the deployed policy reacts to the force-utilization observation.
In this ablation, we show that this observation is necessary for the behavior rather than incidental to it.
To isolate its contribution, we train a second embankment checkpoint that is identical except that the force-utilization term is removed from the observation, and evaluate both checkpoints over 500 simulated rollouts each.
The raw force-normalization value $F_{\max}$ is fixed at the highest machine-capability value observed by either policy during training, so both policies experience negative reward for exceeding this limit. 
The two rollout sets differ only in whether the policy observes its currently applied force during training and test. 
Figure~\ref{fig:fortification_force_ablation_force_usage} compares peak interaction force per rollout, normalized by $F_{\max}$, between the two checkpoints. 
The force-aware policy concentrates peak force closer to the machine-capability line with only few exceeding samples. 
The policy lacking the force observation is more conservative on average and more variable, with a tail of rollouts exceeding machine capability. 
Without the force observation, the policy cannot sense proximity to the limit during the stroke and back off in time, so it either stops short of the available capability or overshoots it. 
The force-aware policy exploits more of the available capability throughout the stroke. 
Median crest-height gain per rollout is \SI{0.75}{\m} (interquartile range \SIrange{0.56}{0.937}{\m}) with the force observation, versus \SI{0.68}{\m} (interquartile range \SIrange{0.5}{0.93}{\m}) without it. 
Removing the force-utilization observation therefore degrades both capability exploitation and embankment height, supporting the force-utilization observation as a necessary observation for successful task execution.

\begin{figure*}
    \centering
    \includegraphics[width=1.0\linewidth]{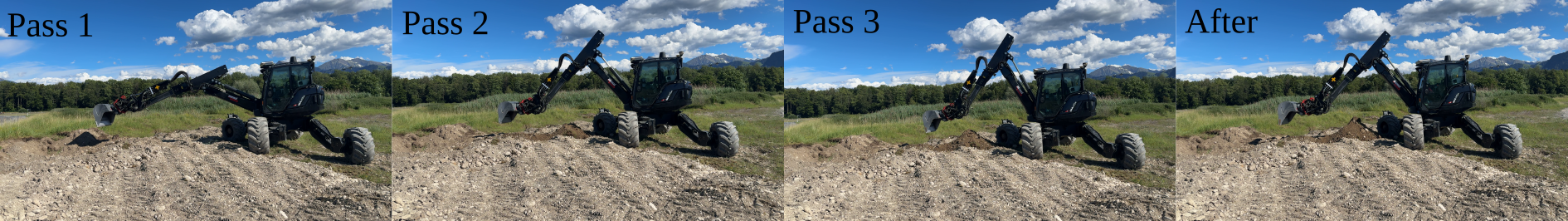}
    \caption{Terrain shape adaptation experiment. We carry out multiple passes in the same workspace. The policy adapts the motion based on the observed terrain shape. Images show the initial setup and the state after the first, second, and third stroke.}
    \label{fig:multi_stroke_observation_mosaic}
\end{figure*}

\begin{figure}
    \centering
    \includegraphics[width=0.8\linewidth]{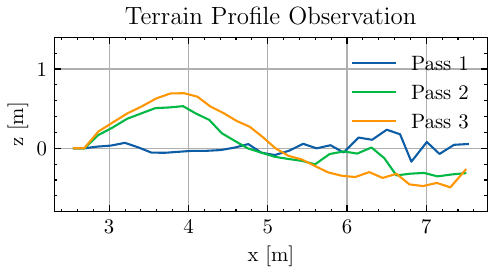}
    \caption{Terrain shape observations for each pass. First pass with flat terrain and second and third with increasing embankment.}
    \label{fig:terrainProfileObservation}
\end{figure}

\begin{figure}
    \centering
    \includegraphics[width=0.8\linewidth]{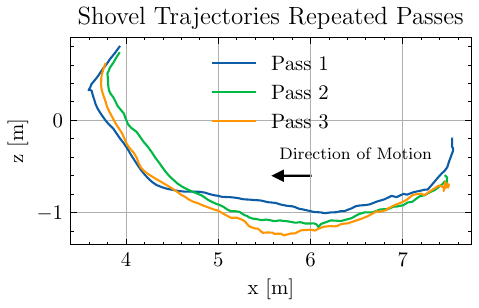}
    \caption{Shovel trajectories for multiple passes on same workspace. The initial pass on flat ground is the most shallow one, with the second and third cutting deeper into the material.}
    \label{fig:eeTrajectoriesMultiplePasses}
\end{figure}

\vq{par:soilProfile}{Does the policy adapt to different soil shapes?}
We validate adaptation to the soil-profile observation by running the embankment policy multiple times at the same location on soft soil.
Figure~\ref{fig:multi_stroke_observation_mosaic} shows the experiment setup with a flat initial surface and an embankment increasing in size on each stroke.
Figure~\ref{fig:terrainProfileObservation} shows the individual terrain shape observations for the three passes.
The tracked trajectories in Figure~\ref{fig:eeTrajectoriesMultiplePasses} show visible adaptation to terrain shape.
The first stroke is the shallowest and reaches furthest back.
The second and third passes become progressively deeper and respect the existing embankment with a less aggressive dumping motion.

Because the experiment soil has little cohesion, material from the sides and from the embankment collapsed back into the trench, which created similar initial conditions and therefore similar second and third strokes.
During embankment construction, this is compensated by selecting overlapping workspaces.


\begin{figure}
    \centering
    \includegraphics[width=1.0\linewidth]{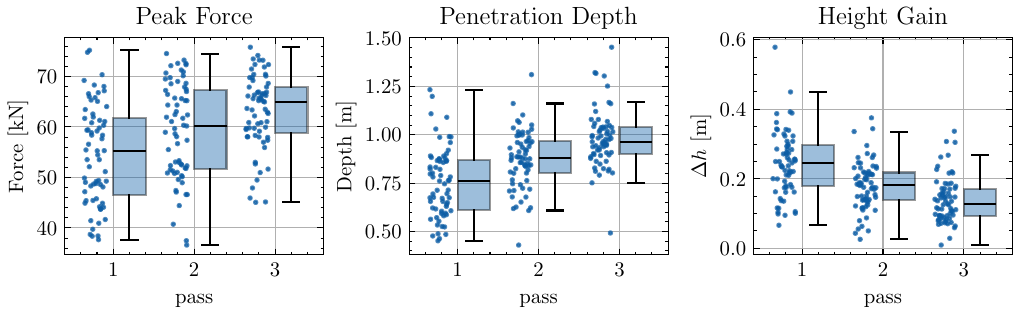}
    \caption{Peak interaction force, penetration depth, and height gain per stroke, grouped by pass number (first, second, third scoop at the same attack point), aggregated over all attack points of the \SI{45}{\min} embankment mission that received at least three scoops. Boxes show the interquartile range and median. Individual strokes are overlaid as points.}
    \label{fig:fortification_per_stroke_stats}
\end{figure}

This result is consistent with statistics recorded across the full \SI{45}{\min} mission, where every attack point receiving multiple scoops provides an independent repetition of the same first/second/third-pass comparison.
Figure~\ref{fig:fortification_per_stroke_stats} reports peak interaction force, penetration depth, and height gain per stroke, grouped by pass number.
Median penetration depth increases monotonically with pass number, from about \SI{0.76}{\m} on the first pass to about \SI{0.88}{\m} on the second and about \SI{0.96}{\m} on the third, while median height gain per stroke falls from about \SI{0.25}{\m} to about \SI{0.18}{\m} and about \SI{0.13}{\m}.
Median peak interaction force rises modestly, from about \SI{56}{\kN} on the first pass to about \SI{65}{\kN} on the third, consistent with the policy scraping into progressively more resistant, less-disturbed material.

% ---------------------------------------------

\vq{par:sim2realExperiments}{Is the soil from the experiments within distribution of what the simulator predicts?}
\begin{figure*}
    \centering
    \includegraphics[width=1.0\linewidth]{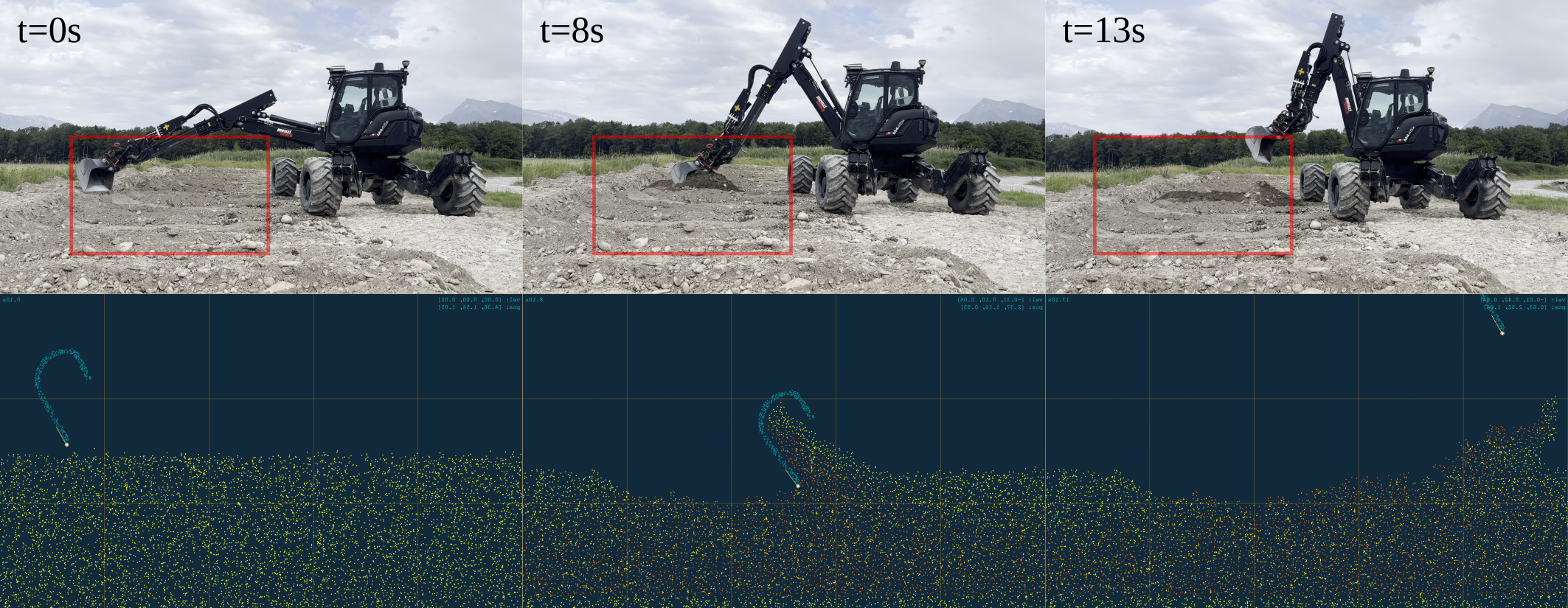}
    \caption{Single replayed scraping motion used to check that the encountered soil lies within the randomized training distribution. The measured M445 shovel pose is replayed in simulation from a laser-scan terrain initialization with soil parameters selected manually inside the training range. Snapshots at 0, 8, and 13~s show conditions before, during, and after the interaction.}
    \label{fig:sim2real_mosaic}
\end{figure*}

This experiment tests that the soil encountered in our field experiments lies within our training distribution.
Because the soil model is overparameterized with respect to a single trajectory, several parameter combinations yield similar force and displacement histories.
We therefore do not solve the inverse problem of inferring soil parameters from measurements.
Soil parameters are hand-picked from the training distribution in order to demonstrate qualitatively, that interactions encountered during real-world experiments have been seen during training.
We use a scraping motion that starts close to the soil, penetrates about \SI{50}{\cm}, and then pulls out of the material as shown in Figure~\ref{fig:sim2real_mosaic}.
On the M445, we log interaction forces and shovel poses and use a laser scan to record the terrain shape after the pass.
In simulation, we kinematically replay the recorded shovel poses on a terrain initialized from the pre-pass laser scan, and log contact forces and the final terrain shape.
Since the simulator is two-dimensional, its contact force is a sectional quantity, and one constant is required to compare it with the measured force of the \SI{1300}{\mm} wide bucket.
This scaling factor is calibrated by matching the magnitude offset of two force profiles of a different motion, sets only the overall force scale, and is not fitted per axis, per sample, or to the temporal shape.

\begin{figure}
    \centering
    \includegraphics[width=0.8\linewidth]{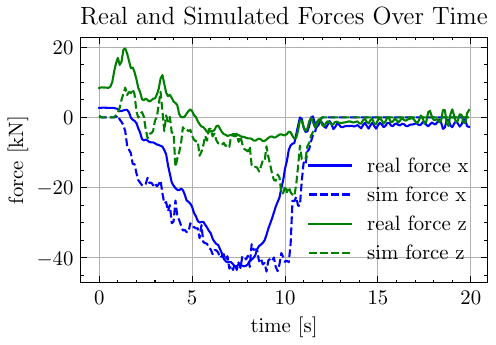}
    \caption{Interaction-force comparison for the single motion shown in Figure~\ref{fig:sim2real_mosaic}. The measured M445 pose is replayed in simulation and the sectional simulated force is scaled by the single constant factor for both components. Force magnitudes and temporal shapes agree for this tested motion.}
    \label{fig:ForceComparison006}
\end{figure}

\begin{figure}
    \centering
    \includegraphics[width=0.8\linewidth]{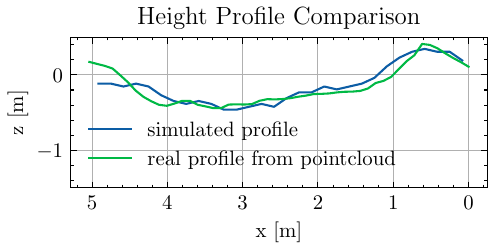}
    \caption{Final terrain shape comparison after the single replayed interaction in Figure~\ref{fig:sim2real_mosaic}. The real profile is computed from a laser-scan point cloud, and the simulated profile is extracted from the particle simulation.}
    \label{fig:HeightComparison006}
\end{figure}

Figure~\ref{fig:ForceComparison006} shows that the x- and z-direction forces agree in magnitude and temporal shape between simulation and reality. 
Because the scale factor is a single constant, this agreement in shape and in the ratio between the two components is a property of the simulated interaction.
The remaining differences are mostly attributable to the residual parameter mismatch and to the idealized perfectly homogeneous soil of the simulator.
Figure~\ref{fig:HeightComparison006} compares the final terrain profile along the motion path.
The simulated profile follows the measured trend. 
The flow and collapsing behavior of the real material is visually matching that in our simulation.
Taken together, the force and profile comparisons indicate that a similar soil behavior was included at training time.

\subsection{Limitations}
Several limitations remain. 
First, the learned policies alone do not solve a long-horizon worksite task.
To keep observations independent of embodiment and low-level controller dynamics, we omit history and use no time-aware network, which keeps observations in-distribution under changing controller dynamics but also prevents the policy from learning multi-stroke motions.
This shortcoming is clearly visible at the compaction skill with successful single stroke compaction, but no re-positioning or pushing from above being learned.
Longer-horizon tasks therefore require embedding these single-stroke skills into a higher level controller.
Similarly, the current reward formulation does not allow free-form shape conditioning.

Second, the normalized interface reduces but does not remove embodiment dependence.
The same checkpoint can be reused only when workspace aspect ratio, perception, force normalization, and low-level tracking preserve the physical meaning of the policy observation and action.
The present experiments use compatible shovel geometries, and the simulator does not train over arbitrary bucket shapes.
Transfer to machines with incompatible buckets, workspaces, or compliant low-level behavior would require additional training.

Third, the simulator is two-dimensional and uses simplifications to maintain reinforcement learning compatible speeds.
While in-plane soil mechanics and force profiles are captured, no heterogeneous materials and three-dimensional effects are seen during training.
These shortcomings are currently compensated by the planner and high-level controller through task repetition and overlapping workspaces.
Furthermore, the simulation replay experiment uses manually selected soil parameters and should be read as qualitative support for the training model.

\section{Conclusion}
We presented a particle-simulation reinforcement-learning formulation for local excavator soil manipulation that learns task-space shovel motions from rewards on simulated material state, including terrain shape, material displacement, contact forces, and compaction memory.
The learned policy is separated from machine-specific actuation through a calibrated perception and low-level motion control interface.

The experiments show material-aware behavior on difficult earthmoving tasks.
Our embankment policy changes its trajectory in response to soft and hard soil, the observed terrain profile, and the indicated machine force capability, and embedded in a worksite state machine it constructs a \SI{42}{\m} embankment under field conditions at a progression speed comparable to an expert operator, with a higher profile and lower height variation.
Backfilling and compaction add qualitative transfer of two more material-state aware policies on both platforms.
The central result is that particle-based reinforcement learning provides a practical local skill layer for contact-rich soil manipulation when the simulator exposes task-relevant material state, and that the calibrated normalized interface is the relevant abstraction for reuse on different machines.

\section{Future Work}
Currently, bucket shape is fixed in simulation. Transfer is therefore shown for compatible bucket geometries, and future work should train over broader tool-geometry distributions.
Future work should add more complex tasks on longer time horizons with a history formulation that does not break low-level controller independence.
In 3D, future policies should learn tasks that are aware of the limitations of the cabin joint and use it similarly to a human operator, for example when closing a trench.

\section*{Acknowledgement}
This work receives funding from Hexagon Innovation Hub GmbH.
A grant from the Swiss National Supercomputing Centre (CSCS) under project ID d130 on Alps supports this work.
The NVIDIA Academic Grant Program supports this research through NVIDIA Warp.

\section*{AI Disclaimer}
The software development leading to this publication has been assisted by generative AI (Copilot, Claude Opus 4.5 and 5.0, Codex).
The authors used Claude 4.5 and 5.0, GPT-5.4 to check grammar and improve language clarity in this manuscript, but did not use generative tools for original research content, data, or conclusions. 
The authors reviewed and edited all generated material and take full responsibility for all content of this publication.

\bibliographystyle{IEEEtran}
\bibliography{sn-bibliography}

\appendices
\newpage

\section{Soil Simulator Implementation}
\label{app:simulator}
In this appendix, we discuss the implementation details of the two-dimensional explicit \ac{MPM} soil simulation.
We implement our solver using NVIDIA Warp, which provides GPU kernels written in plain Python.
Relative to a standard \ac{MPM} solver with a Drucker-Prager update, the implementation adds, besides the compaction memory and the shovel coupling, compaction-dependent hardening and a modified elastic volume response with gradual reference-volume shrinkage.

\paragraph{Simulation setup}

Our computational domain has size
  \(
  L_x \times L_z = 5.0 \,\mathrm{m} \times 3.0 \,\mathrm{m}
  \)
  and is discretized by a Cartesian background grid with
  \(
  N_x \times N_z = 80 \times 48
  \)
  cells, resulting in grid spacing
  \begin{equation}
  \Delta x = \frac{L_x}{N_x}.
  \end{equation}
  A physics time step of
  \(
  \Delta t = 2\times 10^{-3}\,\mathrm{s}
  \) is used, which provides a good tradeoff between accuracy and performance while maintaining stability.
  The controller runs at \SI{10}{\Hz}, i.e., with the control period \(\Delta t_c=0.1\,\mathrm{s}\), thus executing \(N_{\mathrm{sub}}=\Delta t_c/\Delta t=50\) physics update steps before each control update.
  Soil is represented by classical material points, while the shovel is represented by a separate set of particles that follow prescribed rigid-body kinematics for straightforward force coupling. 
  The solver advances the system with the standard MPM sequence
  \[
  \text{clear grid} \rightarrow \text{P2G} \rightarrow \text{grid update} \rightarrow \text{G2P}.
  \]

  \paragraph{Particle and grid state}
  Each soil particle \(p\) stores position \(\mathbf{x}_p \in \mathbb{R}^2\), velocity \(\mathbf{v}_p \in \mathbb{R}^2\), deformation gradient \(\mathbf{F}_p \in \mathbb{R}^{2\times
  2}\), affine velocity matrix \(\mathbf{C}_p \in \mathbb{R}^{2\times 2}\), a reference volume \(V_p\), and a scalar plastic volumetric compaction memory \(\nu_p\). The grid stores nodal mass
  and momentum separately for soil and shovel:
  \[
  (m_i^{\mathrm{soil}}, \mathbf{p}_i^{\mathrm{soil}}), \qquad
  (m_i^{\mathrm{shovel}}, \mathbf{p}_i^{\mathrm{shovel}}).
  \]
  We use the affine term in APIC form. 
  In addition, a one-dimensional column accumulator stores mass-weighted compaction-memory values for reward evaluation of some learned tasks.

  \paragraph{Interpolation}
  We use quadratic B-spline interpolation with a \(3\times 3\) stencil. Let
  \(
  \mathbf{x}_p / \Delta x
  \)
  be the particle position in grid coordinates, and let the lower-left stencil corner be
  \[
  \mathbf{b}_p = \left\lfloor \frac{\mathbf{x}_p}{\Delta x} - \frac{1}{2} \right\rfloor.
  \]
  For the local coordinate
  \(
  \boldsymbol{\xi}_p = \mathbf{x}_p/\Delta x - \mathbf{b}_p
  \),
  the one-dimensional weights are
  \begin{align}
  w_0(\xi) &= \frac{1}{2}(1.5-\xi)^2, \\
  w_1(\xi) &= 0.75-(\xi-1)^2, \\
  w_2(\xi) &= \frac{1}{2}(\xi-0.5)^2,
  \end{align}
  and the two-dimensional weight for grid node \(i,j\) is
  \begin{equation}
  w_{p,ij} = w_i(\xi_{p,x})\, w_j(\xi_{p,z}).
  \end{equation}

  \paragraph{MPM transfer and kinematics}
  For soil particles, the deformation gradient is updated explicitly as
  \begin{equation}
  \mathbf{F}_p^{n+1,*} = \left( \mathbf{I} + \Delta t\, \mathbf{C}_p^n \right)\mathbf{F}_p^n.
  \label{eq:F_update}
  \end{equation}
  The P2G transfer uses APIC momentum:
  \begin{align}
  \mathbf{p}_i^{\mathrm{soil}} &\mathrel{+}= w_{pi}
  \left(
  m_p \mathbf{v}_p + \mathbf{A}_p (\mathbf{x}_i - \mathbf{x}_p)
  \right),
  \\
  m_i^{\mathrm{soil}} &\mathrel{+}= w_{pi} m_p,
  \label{eq:p2g}
  \end{align}
  where
  \begin{equation}
  \mathbf{A}_p = \mathbf{P}_p + m_p \mathbf{C}_p
  \end{equation}
  contains both stress and affine velocity contributions. After grid operations, G2P reconstructs velocity $\mathbf{v}_p^{n+1}$ and affine velocity $\mathbf{C}_p^{n+1}$ as
  \begin{align}
  \mathbf{v}_p^{n+1} &= \sum_i w_{pi}\, \mathbf{v}_i^{n+1}, \\
  \mathbf{C}_p^{n+1} &= 4 \Delta x^{-1} \sum_i w_{pi}\, \mathbf{v}_i^{n+1} \otimes
  \left(\mathbf{x}_i/\Delta x - \mathbf{x}_p/\Delta x\right).
  \label{eq:g2p}
  \end{align}
  Particle positions are then updated by
  \begin{equation}
  \mathbf{x}_p^{n+1} = \mathbf{x}_p^n + \Delta t\, \mathbf{v}_p^{n+1}.
  \end{equation}

  \paragraph{Elastic predictor and Drucker-Prager plasticity}
  We compute the elastic response with a corotated model. We perform a polar decomposition
  \begin{equation}
  \mathbf{F} = \mathbf{R}\mathbf{U},
  \end{equation}
  where \(\mathbf{R}\) is rotation and \(\mathbf{U}\) is the symmetric stretch tensor. Let \(s_1,s_2\) be the principal stretches of \(\mathbf{U}\), and define logarithmic strains
  \begin{equation}
  \boldsymbol{\varepsilon} =
  \begin{bmatrix}
  \log s_1 \\
  \log s_2
  \end{bmatrix}.
  \end{equation}
  The deviatoric part is
  \begin{equation}
  \hat{\boldsymbol{\varepsilon}}
  =
  \boldsymbol{\varepsilon}
  -
  \frac{1}{2}\operatorname{tr}(\boldsymbol{\varepsilon})
  \begin{bmatrix}
  1 \\ 1
  \end{bmatrix}.
  \end{equation}
  With Lam\'e parameters \(\mu,\lambda\), friction angle \(\phi\), and cohesion \(c\), we use
  \begin{align}
  \alpha &= \sqrt{\frac{2}{3}} \, \frac{2 \sin\phi}{2-\sin\phi}, \\
  k_c &= \frac{4 c \cos\phi}{\sqrt{3}(2-\sin\phi)}.
  \end{align}
  Both expressions adapt the circumscribed Drucker-Prager fit to the Mohr-Coulomb criterion, whose three-dimensional form carries \((3-\sin\phi)\) in the denominator~\cite{klar2016drucker}, to our two-dimensional solver by replacing \(3\) with \(2\), consistent with the \(\tfrac{1}{2}\operatorname{tr}(\boldsymbol{\varepsilon})\) deviatoric split used in place of \(\tfrac{1}{3}\operatorname{tr}(\boldsymbol{\varepsilon})\).
  With elasticity the deviatoric Kirchhoff stress is \(2\mu\hat{\boldsymbol{\varepsilon}}\) and \(\operatorname{tr}(\boldsymbol{\tau})=(d\lambda+2\mu)\operatorname{tr}(\boldsymbol{\varepsilon})\), so dividing the stress-space criterion by \(2\mu\) yields a dimensionless measure in which the volumetric term carries \((d\lambda+2\mu)/(2\mu)\) and the cohesive offset becomes \(k_c/(2\mu)\).
  We further cap the frictional contribution, giving the capped Drucker-Prager measure~\eqref{eq:dp_yield} from the main paper.

  \paragraph{Plastic volumetric compaction memory}
  The implementation augments the Drucker-Prager update with a persistent scalar plastic volumetric compaction memory \(\nu_p\). 
  This memory records mostly hydrostatic plastic compression and later supports compaction-related material hardening, volumetric stress response, volume shrinkage and can be used for the task reward.
  The magnitude of the elastic logarithmic strain is converted into a capped loading factor
  \begin{equation}
  r_p =
  \min\!\left(r_p^{\max},\,
  \frac{\|\boldsymbol{\varepsilon}\|}{\varepsilon_{\mathrm{th}}}
  \right),
  \label{eq:rate_factor}
  \end{equation}
  where \(\varepsilon_{\mathrm{th}}\) is a strain threshold and \(r_p^{\max}\) caps the loading factor.
  Heavier elastic loading accumulates compaction memory faster, and the cap limits the contribution of a single step.
  Expressing the factor as a strain ratio keeps it dimensionless and independent of \(\Delta t\). Let
  \begin{equation}
  e_c = \max\!\left(0,-\operatorname{tr}(\boldsymbol{\varepsilon})\right),
  \qquad
  \rho_h =
  \frac{e_c}{e_c+\|\hat{\boldsymbol{\varepsilon}}\|+\epsilon_{\mathrm{reg}}}
  \end{equation}
  denote the compressive logarithmic volumetric strain and a hydrostatic-compression ratio, where \(\epsilon_{\mathrm{reg}}\) is a small constant that avoids division by zero when both terms vanish.
  Memory accumulation is triggered only when compression dominates shear:
  \begin{equation}
  e_c > \max(e_c^{\min}, k_c \eta_y),
  \qquad
  \rho_h > \rho_h^{\min},
  \end{equation}
  where \(\eta_y\) converts the cohesive strength into a strain threshold and therefore carries units of inverse stress.
  The constant floor \(e_c^{\min}\) retains a minimum compression requirement for materials with little cohesion, independent of \(k_c\eta_y\).
  This threshold governs the compaction memory only and does not enter the stress response.
  The hydrostatic-ratio threshold \(\rho_h^{\min}\) requires compression to dominate the local strain state without demanding purely volumetric loading, since shovel loading is typically a mix of compression and shear.
  In that case the scalar memory is incremented by
  \begin{align}
  \Delta \nu_p &=
  \left(e_c-\max(e_c^{\min},k_c\eta_y)\right)\rho_h
  \min(r_p,r_p^{\mathrm{comp}}),
  \\
  \nu_p &\leftarrow \min(\nu_p+\Delta\nu_p,\nu_{\max}).
  \label{eq:compaction_update}
  \end{align}
  Thus heavier elastic loading can increase how quickly compaction memory accumulates, up to the additional cap \(r_p^{\mathrm{comp}}\), but the elastic Drucker-Prager return mapping remains responsible for the stress-producing deformation gradient.

  \paragraph{Stress computation}
  Before computing stress, the compaction memory hardens the material parameters. With \(\nu_p^{h}=\min(\nu_p,\nu_{\max})\), we use
  \begin{align}
  h_p &= 1 + k_h \nu_p^{h}, \\
  \mu_p &= h_p \mu_0,\qquad
  \lambda_p = h_p \lambda_0,\qquad
  c_p = h_p c, \\
  \phi_p &= \phi + \min(k_\phi\nu_p^{h},\Delta\phi_{\max}),
  \end{align}
  where \(k_h\) is the stiffness-hardening coefficient, \(k_\phi\) is the friction-angle hardening rate, and \(\Delta\phi_{\max}\) caps the total friction-angle increase from compaction hardening.
  Section~\ref{sub:solverDetails} summarizes this as a common hardening factor \(1+k_h\nu_p\); precisely, stiffness and cohesion scale with \(h_p\), whereas the friction angle increases additively and is capped.

  After the constitutive update, we use the Cauchy-like stress term
  \begin{align}
  \boldsymbol{\sigma}_p
  &=
  2\mu_p (\mathbf{F}_p - \mathbf{R})\mathbf{F}_p^{\mathsf{T}}
  +
  \lambda_p J_p^{e} (J_p^{e}-1)\mathbf{I},
\\
  J_p^{e} &= \min\!\left(1, \det(\mathbf{F}_p)\exp(\nu_p)\right).
  \label{eq:stress}
  \end{align}
  The modified volume ratio \(J_p^{e}\) treats the compaction memory as isotropic plastic volume loss.
  It can relax compressive rebound, but it is clamped so that compaction memory does not create artificial tensile suction.
  This is converted into the affine stress contribution for P2G as
  \begin{equation}
  \mathbf{P}_p
  =
  -
  \Delta t \, V_p \, \kappa_c \, 4\Delta x^{-2}\, \boldsymbol{\sigma}_p,
  \label{eq:stress_affine}
  \end{equation}
  where \(\kappa_c\) is a near-unity compressibility correction factor. 
  This extra scalar factor is another pragmatic modification to tune bulk response without changing the basic constitutive form.

  During G2P, the local grid mass gives an observed volume
  \begin{equation}
  V_p^{\mathrm{obs}} = \Delta x^2 \frac{m_p}{\sum_i w_{pi}m_i}.
  \end{equation}
  If compaction memory is present and \(V_p^{\mathrm{obs}}\) is sufficiently smaller than the current reference volume set at reset time, \(V_p\) is reduced gradually toward the observed packed volume with a per-step shrinkage limit. 
  Therefore, compacted soil occupies less volume in subsequent steps.

  \paragraph{Rigid shovel representation and soil interaction}
  We represent stiff objects as a set of particles attached to a prescribed rigid transform, and resolve contact with separate soil and shovel velocity fields on the grid, following the multi-velocity-field \ac{MPM} contact approach of~\cite{bardenhagen2000granular}. 
  Let the shovel pose be
  \(
  (\mathbf{x}_s,\theta_s)
  \)
  with translational velocity \(\mathbf{u}_s\) and angular velocity \(\omega_s\). 
  For a shovel particle at offset \(\mathbf{r}_p\) from the shovel origin, we assign the particle velocity
  \begin{equation}
  \mathbf{v}_p^{\mathrm{shovel}}
  =
  \mathbf{u}_s
  +
  \omega_s
  \begin{bmatrix}
  r_{p,z} \\
  -r_{p,x}
  \end{bmatrix}.
  \label{eq:rigid_particle_velocity}
  \end{equation}
  We transfer these shovel particles to a separate solid grid:
  \begin{align}
  \mathbf{p}_i^{\mathrm{shovel}} &\mathrel{+}= w_{pi}\, m_p \mathbf{v}_p^{\mathrm{shovel}}, \\
  m_i^{\mathrm{shovel}} &\mathrel{+}= w_{pi}\, m_p.
  \end{align}
  In addition, the solver accumulates the unweighted particle gradient
  \begin{equation}
  \mathbf{g}_i \mathrel{+}= \mathbf{x}_i - \mathbf{x}_p,
  \end{equation}
  which serves as a local estimate of contact normal direction,
  \begin{equation}
  \mathbf{n}_i = \frac{\mathbf{g}_i}{\|\mathbf{g}_i\|}.
  \end{equation}
  The normal is estimated directly from the rasterized shovel particle cloud for higher performance on parallelized compute.

  Contact is resolved on the grid after converting soil momentum to velocity and adding gravity. 
  For a node containing both soil and shovel particles, we compute
  \begin{align}
  \mathbf{v}_i &=
  \frac{\mathbf{p}_i^{\mathrm{soil}}}{m_i^{\mathrm{soil}}}
  + \Delta t\,\mathbf{g},
    \\
  \mathbf{u}_i &=
  \frac{\mathbf{p}_i^{\mathrm{shovel}}}{m_i^{\mathrm{shovel}}},
  \\
  \mathbf{v}_{\mathrm{rel},i} &= \mathbf{v}_i-\mathbf{u}_i,
  \end{align}
  and detect contact motion into the shovel if
  \begin{equation}
  \chi_i = \mathbf{v}_{\mathrm{rel},i} \cdot (-\mathbf{n}_i) > 0.
  \label{eq:contact_closing_speed}
  \end{equation}
  Here \(\chi_i\) is the normal closing speed at the grid node.
  The normal component of the relative velocity is removed by the collision velocity increment
  \begin{equation}
  \Delta \mathbf{v}_{n,i}
  =
  \chi_i \mathbf{n}_i.
  \label{eq:normal_contact}
  \end{equation}
  Our solver additionally applies a Coulomb-like tangential friction. With the tangential relative velocity and its direction,
  \begin{equation}
  \mathbf{v}_{t,i}
  =
  \mathbf{v}_{\mathrm{rel},i}
  -
  (\mathbf{v}_{\mathrm{rel},i}\cdot \mathbf{n}_i)\mathbf{n}_i,
  \qquad
  \hat{\mathbf{t}}_i
  =
  \frac{\mathbf{v}_{t,i}}{\|\mathbf{v}_{t,i}\|},
  \end{equation}
  and the clamped friction speed
  \begin{equation}
  \gamma_i
  =
  \min(\mu_c \chi_i,\|\mathbf{v}_{t,i}\|),
  \label{eq:friction_speed}
  \end{equation}
  the tangential velocity increment is
  \begin{equation}
  \Delta \mathbf{v}_{t,i}
  =
  -\gamma_i\hat{\mathbf{t}}_i,
  \label{eq:friction_contact}
  \end{equation}
  where \(\mu_c\) is the shovel-soil friction coefficient and \(\Delta \mathbf{v}_{t,i}=\mathbf{0}\) for \(\|\mathbf{v}_{t,i}\|=0\).
  The clamp prevents friction from reversing the available tangential slip in a single grid update.
  Both increments are applied to the grid velocity,
  \begin{equation}
  \mathbf{v}_i
  \leftarrow
  \mathbf{v}_i
  +
  \Delta \mathbf{v}_{n,i}
  +
  \Delta \mathbf{v}_{t,i},
  \label{eq:contact_velocity_update}
  \end{equation}
  which is equivalent to applying the impulse
  \begin{equation}
  \mathbf{J}_{i}
  =
  m_i^{\mathrm{soil}} \left(\Delta \mathbf{v}_{n,i}+\Delta \mathbf{v}_{t,i}\right)
  =
  m_i^{\mathrm{soil}}\left(\chi_i\mathbf{n}_i-\gamma_i\hat{\mathbf{t}}_i\right)
  \label{eq:contact_impulse}
  \end{equation}
  to the soil.
  We obtain the force accumulator used by the policy by dividing this impulse by the MPM step size and summing over all contact nodes and all \(N_{\mathrm{sub}}\) physics substeps in one control step:
  \begin{equation}
  \mathbf{F}^{s}_t
  =
  \frac{1}{N_{\mathrm{sub}}}
  \sum_{\ell=1}^{N_{\mathrm{sub}}}
  \sum_{i\in\mathcal{C}_{t,\ell}}
  \frac{m_{i,\ell}^{\mathrm{soil}}}{\Delta t}
  \left(\chi_{i,\ell}\mathbf{n}_{i,\ell}-\gamma_{i,\ell}\hat{\mathbf{t}}_{i,\ell}\right).
  \label{eq:shovel_contact_force}
  \end{equation}
  The contact set \(\mathcal{C}_{t,\ell}\) contains grid nodes with nonzero soil mass, nonzero shovel mass, a valid rasterized normal, and positive closing speed.
  In our solver this is the quantity stored as the shovel collision force.
  Because the accumulator contains the normal and the frictional term, it also represents the tangential resistance that dominates scraping and cutting motions.
  The force therefore measures the effort that the commanded shovel motion would require from the machine, and we use it for the force observation and reward penalties.
  On the real machine reaction is carried by the excavator through its hydraulics and chassis into the ground rather than accelerating the tool, so the machine acts as a momentum sink on the time scale of one interaction.

  \paragraph{Boundary conditions}
  The bottom few grid rows are treated as sticky
  \(
  \mathbf{v}_i=\mathbf{0}
  \),
  while the remaining outer boundary suppresses only outward normal motion. These boundary conditions approximate the ground below, where force is distributed as inter-particle shear.

  \paragraph{Parallelization}
  The implementation is batched over environments.
  All particle and grid arrays carry the environment index in their leading dimension, and every kernel processes all environments
  simultaneously. 
  The main kernels parallelize additionally either over particles or over grid nodes
  \begin{itemize}
  \item \emph{particle-parallel kernels}: shovel particle velocity update, P2G, and G2P;
  \item \emph{grid-parallel kernels}: grid clearing and grid update/contact.
  \end{itemize}
  Thus, for \(N_{\mathrm{env}}\) parallel simulations with \(N_p\) particles and \(N_g=N_xN_z\) grid nodes per environment, the dominant work scales as
  \begin{align}
  &\mathcal{O}\!\left(N_{\mathrm{env}} N_p\right)
  \quad \text{for particle kernels,}
  \\
  &\mathcal{O}\!\left(N_{\mathrm{env}} N_g\right)
  \quad \text{for grid kernels.}
  \end{align}
  We capture the full sequence of physics substeps between two control updates once as a Warp execution graph and replay it during training for reduced launch overhead. 
  The slowest kernels are P2G and G2P, which contain the major grid scattering and gathering operations.
  For the P2G kernel on GH200, memory access accounts for about 80\% of the runtime and soil-model computation for about 20\%.
  All other kernels are more than an order of magnitude faster.
  We experimented with several speed improvements, including switching to gathering, custom data structures, and cache optimization. 
  None significantly improved the runtime of the P2G kernel, so we kept the straightforward scattering implementation.

\section{Learning Formulation}
\label{app:learning}
This appendix details the environment initialization, randomization, rewards, and termination conditions summarized in Sections~\ref{sub:learningSoilInteraction}--\ref{sub:termination}.

\subsection{Environment Initialization and Randomization}
\label{app:randomization}
\paragraph{Initialization}
Terrain initialization is sampled at reset time from precomputed particle distributions.
We generate terrain shapes by randomly sampling positive and negative features along the workspace with subsequent smoothing for physical viability.
\begin{equation}
h_0(x)
= b + \left(K_{\mathrm{Hann}} *
\sum_{j=1}^{n_b} \alpha_j
\mathbb{I}\left[|x-c_j|\leq \frac{w_j}{2}\right]\right)(x),
\label{eq:precomputed_profile}
\end{equation}
where \(b\) is the baseline height, \(K_{\mathrm{Hann}}\) is a normalized Hann smoothing kernel, \(n_b\) is the number of sampled features, and each feature is parameterized by center \(c_j\), width \(w_j\), and signed height \(\alpha_j\).
Positive features \((\alpha_j>0)\) represent local piles or embankments, while negative features \((\alpha_j<0)\) represent pits, trenches, or missing material.
Particles are then sampled uniformly under \(h_0(x)\), producing a particle set
\begin{equation}
\mathcal{P}_k^0 = \{\mathbf{x}_{p,k}^0\}_{p=1}^{N_p},
\qquad
0 \leq z_{p,k}^0 \leq h_{0,k}(x_{p,k}^0),
\label{eq:precomputed_particles}
\end{equation}
for each precomputed setup \(k\) drawn as $k \sim \mathcal{U}\{1,\ldots,N_{\mathrm{setup}}\}$.

\paragraph{Randomization}
We initialize the shovel pose from task-specific bounds,
\begin{align}
\mathbf{p}^{s}_0
&=\bar{\mathbf{p}}^{s}_0+\boldsymbol{\epsilon}_p,
&
\epsilon^i_p
&\sim \mathcal{U}[-\Delta p_i,\Delta p_i],
\nonumber\\
\theta^s_0
&=\bar{\theta}^s_0+\epsilon_{\theta},
&
\epsilon_{\theta}
&\sim\mathcal{U}[-\Delta\theta,\Delta\theta].
\label{eq:spawn_randomization}
\end{align}
We also sample the machine force capability and shovel speed limit per environment,
\begin{align}
F_{\max}
&\sim\mathcal{U}[F_{\min},F_{\max}^{\mathrm{cfg}}],
&
v_{\max}
&\sim\mathcal{U}[v_{\min},v_{\max}^{\mathrm{cfg}}].
\label{eq:machine_randomization}
\end{align}
The soil parameters are randomized across silt, sand, and clay with different water contents and compaction states.
Achieved soils are classified through manual observation of the interaction and collapse behavior.
Randomization values for the soil parameters are hand-tuned, as specific tasks such as compaction do not work in near-liquid environments.

\subsection{Reward Functions}
\label{app:rewards}
Training uses a two-stage curriculum indexed by \(\ell\).
In the first stage, \(\ell=0\), the curriculum-specific height offset and additional spawn jitter are disabled, and the reward contains an initialization term that keeps the commanded control point near a reference pose.
After the mean return of completed episodes reaches a task-specific threshold \(R_{\mathrm{curr}}\), the environments switch to \(\ell=1\),
\begin{equation}
\ell \leftarrow 1
\quad\text{if}\quad
\bar{R}_{\mathrm{episode}} \geq R_{\mathrm{curr}}.
\end{equation}
The second stage removes this initialization shaping and enables a sampled vertical height offset in the terrain and shovel observations, together with more shovel position spawn jitter.
Several components are shared between tasks:
\begin{align}
r_{\mathrm{act}} &= -\lambda_a \|\mathbf{a}_t\|_2^2,
\label{eq:reward_action_penalty}\\
r_{\mathrm{track}} &=
-\lambda_q\,
\mathbb{I}\left[\|\mathbf{q}^{c}_t-\mathbf{q}^{s}_t\|_2>\delta_q\right]
\|\mathbf{q}^{c}_t-\mathbf{q}^{s}_t\|_2,
\label{eq:reward_tracking_penalty}\\
r_{\mathrm{init}} &=
-\lambda_0\,\mathbb{I}[\ell=0]\,
\|\mathbf{p}^{c}_t-\mathbf{p}^{\mathrm{ref}}\|_2^2.
\label{eq:reward_initialization_penalty}
\end{align}
The action penalty in \eqref{eq:reward_action_penalty} discourages unnecessarily large command increments.
The tracking penalty in \eqref{eq:reward_tracking_penalty} penalizes targets that move too far away from the simulated shovel pose.
The initialization penalty in \eqref{eq:reward_initialization_penalty} is active only for \(\ell=0\) and provides extra gradient for the shovel moving in the right direction.
Following Section~\ref{sub:rewardFunctions}, \(r_{\mathrm{sh}}=r_{\mathrm{act}}+r_{\mathrm{track}}+r_{\mathrm{init}}+r_{\mathrm{term}}\) collects these shared terms together with the task-specific terminal reward \(r_{\mathrm{term}}\) of \Cref{app:termination}.
Reward weights \(\lambda\) and thresholds are task-specific constants, so the same symbol can take different values in different tasks.

\paragraph{Embankment formation}
For embankment formation, each reset stores a target profile \(\mathbf{h}^{\star}\), an attention mask \(\mathbf{m}\in\mathbb{R}_{\geq0}^{N_h}\), and a target coordinate \(x^{\star}\).
The target coordinate is the x coordinate of the center of the desired embankment.
For embankment formation the mask is binary, whereas the backfilling task uses graded weights.
The target feature height \(\mathbf{h}^{\star}\) is sampled above the height normally reachable in one stroke.
Therefore even with perfect execution, the policy cannot move the post stroke particle profile above the target profile, creating a negative learning signal for perfect execution.
The policy observes \(x^{\star}\), but it does not observe \(\mathbf{h}^{\star}\) or \(\mathbf{m}\).
Let \(\mathbf{h}_t\in\mathbb{R}^{N_h}\) be the current sampled terrain profile.
The attention-weighted distance to the embankment target is
\begin{equation}
d_t =
\frac{\sum_{i=1}^{N_h}\mathbb{I}[m_i>m_{\min}]\,m_i\left|h_{t,i}-h^{\star}_{i}\right|}
{\sum_{i=1}^{N_h}\mathbb{I}[m_i>m_{\min}]\,m_i},
\label{eq:height_distance}
\end{equation}
so samples outside the attention zone are ignored and samples inside it contribute proportionally to their weight, with \(m_{\min}=0.1\).
We denote the best profile distance observed so far by
\begin{equation}
d^{\mathrm{best}}_t=\min_{\tau\leq t} d_{\tau}.
\end{equation}
In addition to the shared rewards, the embankment task rewards improvement in this distance and gives a small dense bonus where the target profile contains the positive embankment:
\begin{align}
r^{\mathrm{emb}}_t
&=
\lambda_{d}\left[d^{\mathrm{best}}_{t-1}-d_t\right]_+ \nonumber\\
&\quad+ \lambda_{h}\sum_{i=1}^{N_h}
\mathbb{I}[h_i^{\star}>h_{\mathrm{emb}}]\,(h_{t,i}-h_{\mathrm{base}})
+ r_F + r_{\mathrm{sh}},
\label{eq:embankment_reward}
\end{align}
with \([x]_+=\max(x,0)\).
The progress term rewards reductions in the best observed height-profile distance.
The dense bonus acts on the raised part of the target, and the shared terms enter through \(r_{\mathrm{sh}}\).

The force term discourages commands that exceed the sampled machine capability,
\begin{align}
r_F
&=
-\lambda_{F}\left(
\mathbb{I}[|F_{t,x}|>1]\,F_{t,x}^{2}+\mathbb{I}[|F_{t,z}|>1]\,F_{t,z}^{2}
\right)
\nonumber\\
&\quad
-\lambda_{F_z}\,\mathbb{I}[|F^{s}_{t,z}|>F_{\mathrm{emb},z}]\,|F^{s}_{t,z}|.
\label{eq:embankment_force_reward}
\end{align}
Here \(\mathbf{F}_t=\mathbf{F}^{s}_t/F_{\max}=(F_{t,x},F_{t,z})\) is the shovel contact force normalized by the sampled machine force limit, so a magnitude of one corresponds to the machine capability, and \(F^{s}_{t,z}\) is the vertical component of the unnormalized shovel contact force.
This reward encodes that the local height profile should move closer to the target trench-and-embankment profile while avoiding excessive force and poorly tracked control commands.

\paragraph{Backfilling}
The backfilling reward is formulated as a surface-height objective on terrains initialized with one positive and one negative feature.
The objective is to move material from the positive feature into the negative feature until the profile approaches a uniform height.
The task reuses the target profile \(\mathbf{h}^{\star}\), the attention mask \(\mathbf{m}\), and the attention-weighted distance \(d_t\) of \eqref{eq:height_distance}.
Here the target is the uniform level \(h^{\star}_i=h_{\mathrm{flat}}\) for all \(i\), and the attention weights are raised over the pile and the depression, so that \(d_t\) is dominated by the two features that must be equalized.
In addition to the distance, a Gaussian kernel
\begin{equation}
G(h;\mu,\sigma)=
\frac{1}{\sigma\sqrt{2\pi}}
\exp\left(-\frac{1}{2}\left(\frac{h-\mu}{\sigma}\right)^2\right)
\end{equation}
provides a dense bonus for samples that already sit at the target level.
With \(\Delta_t=d_{t-1}-d_t\) and \([x]_-=\max(-x,0)\), the per-step reward is
\begin{align}
r^{\mathrm{back}}_t
&=
\lambda^{+}_{d}\left[\Delta_t\right]_+
-\lambda^{-}_{d}\left[\Delta_t\right]_-
\nonumber\\
&\quad+\lambda_{G}\sum_{i=1}^{N_h} G(h_{t,i};h^{\star}_i,\sigma_h)
+ r_{\mathrm{trans}}
+ r_{\mathrm{sh}}.
\label{eq:backfilling_reward}
\end{align}
Unlike embankment formation, progress is measured between consecutive steps instead of against the best distance observed so far, so that material pushed away from the target is penalized as well. 
Asymmetric weights keep improvement more valuable than an equally large regression.

A flat profile can in principle also be approached by locally leveling the pile, without filling the depression.
We therefore add a transport term that rewards commanded motion directed from the surplus region towards the deficit region while the shovel is working the material.
Let \(\rho_{t,i}=h_{t,i}-h^{\star}_i\) be the profile residual, \(x_i\) the abscissa of sample \(i\), and \(\epsilon_{\rho}\) a deadband.
The attention-weighted surplus and deficit masses and their centroids are
\begin{align}
w^{\pm}_{t,i}&=\mathbb{I}[m_i>m_{\min}]\,m_i\left[\pm\rho_{t,i}-\epsilon_{\rho}\right]_+,
\nonumber\\
\bar{x}^{\pm}_t&=\frac{\sum_i w^{\pm}_{t,i}\,x_i}{\sum_i w^{\pm}_{t,i}}.
\end{align}
The term is active only when both a surplus and a deficit exist.
Its magnitude is the commanded horizontal displacement projected onto the transport direction and saturated at \(\delta_x\),
\begin{equation}
\Delta x_t=
\mathrm{clip}\!\left(
\mathrm{sign}(\bar{x}^{-}_t-\bar{x}^{+}_t)\,(x^{c}_t-x^{c}_{t-1}),\,
-\delta_x,\,\delta_x
\right),
\end{equation}
and it is gated by three factors that require the shovel to be at the source, at working height, and in contact with the soil,
\begin{align}
g_{\mathrm{src}}&=\left[1-\left|x^s_t-\bar{x}^{+}_t\right|/d_{\mathrm{src}}\right]_+,
\nonumber\\
g_{h}&=\left[1-\left|z^s_t-(h_{\mathrm{flat}}+\Delta h_w)\right|/d_{h}\right]_+,
\nonumber\\
g_{F}&=\min\left(1,\|\mathbf{F}_t\|_2\right),
\quad
g_{w}=\max\left(g_F,\,\beta_w\,g_h\right),
\end{align}
so that
\begin{equation}
r_{\mathrm{trans}}
=
\begin{cases}
\lambda^{+}_{x}\,g_{\mathrm{src}}\,g_{w}\,\dfrac{\Delta x_t}{\delta_x}, & \Delta x_t>0,\\[2mm]
\lambda^{-}_{x}\,g_{\mathrm{src}}\,g_{w}\,\dfrac{\Delta x_t}{\delta_x}, & \Delta x_t\leq 0.
\end{cases}
\label{eq:backfilling_transport_reward}
\end{equation}
Motion that transports material is rewarded when the shovel is loaded, whereas the same motion in free space is worth at most \(\beta_w\) of that value.

\paragraph{Compaction}
The simulator accumulates the compaction memory \(\nu_p\) in the task attention zone \(\mathcal{A}\) and reports its mass-weighted mean
\begin{equation}
\bar{\nu}_t =
\frac{\sum_{i\in \mathcal{A}} \sum_p w_{pi} m_p \nu_{p,t}}
{\sum_{i\in \mathcal{A}} \sum_p w_{pi} m_p},
\label{eq:reward_compaction_state}
\end{equation}
where \(w_{pi}\) are the particle-to-grid interpolation weights.
Let
\begin{equation}
\bar{\nu}^{\mathrm{best}}_t=\max_{\tau\leq t}\bar{\nu}_{\tau}
\end{equation}
after an initial settling period of \(T_{\mathrm{settle}}=2\,\mathrm{s}\), and let \(N^{\mathcal{A}}_t\) be the number of particles in the attention zone.
The compaction reward is
\begin{align}
r^{\mathrm{comp}}_t
&=
\lambda_{p}\left[\bar{\nu}_t-\bar{\nu}^{\mathrm{best}}_{t-1}\right]_+
+\lambda_{\nu}\bar{\nu}_t
-\lambda_{F}\left[\|\mathbf{F}_t\|_2-1\right]_+^2
\nonumber\\
&\quad
+\lambda_{N}\,\min(0,N^{\mathcal{A}}_t-N^{\mathcal{A}}_{t-1})
+ r_{\theta} - \lambda_{t} + r_{\mathrm{sh}},
\label{eq:compaction_reward}
\end{align}
with the orientation penalty \(r_{\theta}=-\lambda_{\theta}\,\mathbb{I}[\theta^c_t>\theta_{\max}\lor\theta^c_t<\theta_{\min}]\).
The reward in \eqref{eq:compaction_reward} consists of a compaction progress term, a dense compaction term, a force-limit penalty, a particle-retention term, an orientation penalty, a time penalty, and the shared terms.
The progress term rewards only improvements over the best compaction observed so far, while the dense term provides a continuous signal once compaction starts.
The orientation penalty keeps the commanded shovel angle in the useful compaction range.
The force-limit penalty penalizes normalized force utilization above the sampled machine capability.
The particle-retention term penalizes particles leaving the attention zone because \(\min(0,N^{\mathcal{A}}_t-N^{\mathcal{A}}_{t-1})\leq0\).
The constant time penalty \(\lambda_{t}\) discourages stalling.

\subsection{Episode Termination}
\label{app:termination}
We compute episode termination and termination reward \(r_{\mathrm{term}}\) at the control rate before evaluating the reward.
All tasks share timeout and workspace-bound checks: if the physics frame count reaches \(N_{\mathrm{step}}\), we mark the episode as a timeout and set \(r_{\mathrm{term}}=0\).
If the shovel leaves the planar workspace,
\begin{equation}
x^s_t < x^{s}_{\min} \quad\lor\quad x^s_t > L_x-\delta_x^{\mathrm{ws}} \quad\lor\quad z^s_t < z^{s}_{\min},
\end{equation}
the episode terminates with a small penalty.
Here \(L_x\) is the workspace length, \(T_{\mathrm{ep}}\) is the fixed episode duration, \(N_{\mathrm{step}}=T_{\mathrm{ep}}/\Delta t\) is the maximum number of physics steps in an episode, and \(x^{s}_{\min}=z^{s}_{\min}\) and \(\delta_x^{\mathrm{ws}}\) are the workspace safety margins.

\paragraph{Embankment formation}
For embankment formation, a successful episode ends when the shovel moves to the upper exit region of the workspace while keeping a shovel angle consistent with finishing the stroke,
\begin{equation}
z^s_t > L_z-\delta_z^{\mathrm{emb}} \quad\land\quad \theta^{\mathrm{emb}}_{\min} < \theta^s_t < \theta^{\mathrm{emb}}_{\max}.
\end{equation}
In this case the terminal reward is the accumulated profile-improvement reward, \(r_{\mathrm{term}}=R^{\mathrm{emb}}_{\mathrm{acc},t}\), with
\begin{equation}
R^{\mathrm{emb}}_{\mathrm{acc},t}
=\sum_{\tau\leq t}
\lambda_{d}
\left[d^{\mathrm{best}}_{\tau-1}-d_{\tau}\right]_+.
\end{equation}
The episode also terminates if the normalized vertical force utilization becomes too negative, \(F_{t,z}<-1\), which prevents the policy from learning strokes that exceed the sampled machine capability with \(r_{\mathrm{term}}=0\).

\paragraph{Backfilling}
For backfilling, the task terminates successfully when the attention-weighted profile distance falls below a tolerance,
\begin{equation}
d_t < \epsilon_d.
\end{equation}
Success gives \(r_{\mathrm{term}}=r^{\mathrm{succ}}_{\mathrm{back}}\).
The episode terminates with \(r_{\mathrm{term}}=r^{\mathrm{bound}}_{\mathrm{back}}\) on a workspace-bound violation, if the shovel angle leaves \(\theta^{\mathrm{back}}_{\min}<\theta^s_t<\theta^{\mathrm{back}}_{\max}\), or if the vertical shovel contact force becomes too negative, \(F^{s}_{t,z}<F^{\min}_{\mathrm{back},z}\).
Timeouts use \(r_{\mathrm{term}}=0\).

\paragraph{Compaction}
The compaction task terminates unsafe motions.
\(\mathbf{F}^{s,\mathrm{loc}}_t=R(-\theta^s_t)\mathbf{F}^{s}_t\) is the shovel contact force expressed in the shovel frame.
In addition to the shared timeout and workspace-bound checks, the episode terminates with \(r_{\mathrm{term}}=r^{\mathrm{fail}}_{\mathrm{comp}}\) if the shovel pulls upward, \(F^s_{t,z}>F^{\mathrm{pull}}_{\mathrm{comp}}\), if the local tangential force is too large, \(|F^{s,\mathrm{loc}}_{t,x}|>F^{\mathrm{tan}}_{\mathrm{comp}}\), or if the local force is dominated by the tangential component instead of the downward compaction component,
\begin{equation}
|F^{s,\mathrm{loc}}_{t,x}| > |F^{s,\mathrm{loc}}_{t,z}| + \Delta F_{\mathrm{comp}}.
\end{equation}
These conditions prevent the policy from satisfying the compaction objective through pulling or cutting.

\section{Field Deployment System}
\label{app:fieldsystem}
This appendix details the worksite supervisor of Section~\ref{sub:stateMachine} and the embankment mission execution of Section~\ref{sub:embankmentExperiment}.

\subsection{Worksite Decomposition and Supervisor}
\label{app:supervisor}
\begin{figure}
\centering
\includegraphics[width=1.0\linewidth]{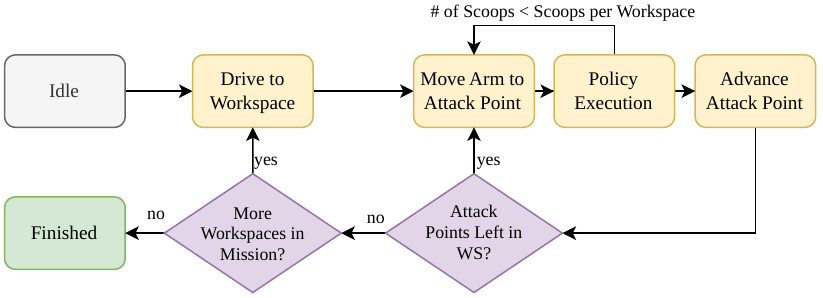}
\caption{State diagram of the supervisory finite state machine.}
\label{fig:supervisoryFiniteState}
\end{figure}

The planner decomposes the task into local workspaces of approximately one shovel width, and selects the next workspace to improve.
At the beginning of the large-scale mission, the planner receives the desired path location and computes a sequence of base poses, aligned workspaces and attack points that cover the geometry.
The reachable portion of the target path from a base pose determines the number of attack points assigned to that pose.
We use a state machine to execute this plan by driving to the next base pose, aligning the cabin with the attack point, capturing an unobstructed height scan and running the learned policy until its terminal condition.
\Cref{app:missionexecution} describes the individual states for the embankment mission.
Criteria for re-execution of the policy can vary based on the task.
Figure~\ref{fig:supervisoryFiniteState} visualizes the supervisory state machine.

\subsection{Embankment Mission Execution}
\label{app:missionexecution}
\begin{figure}
\centering
\includegraphics[width=0.8\linewidth]{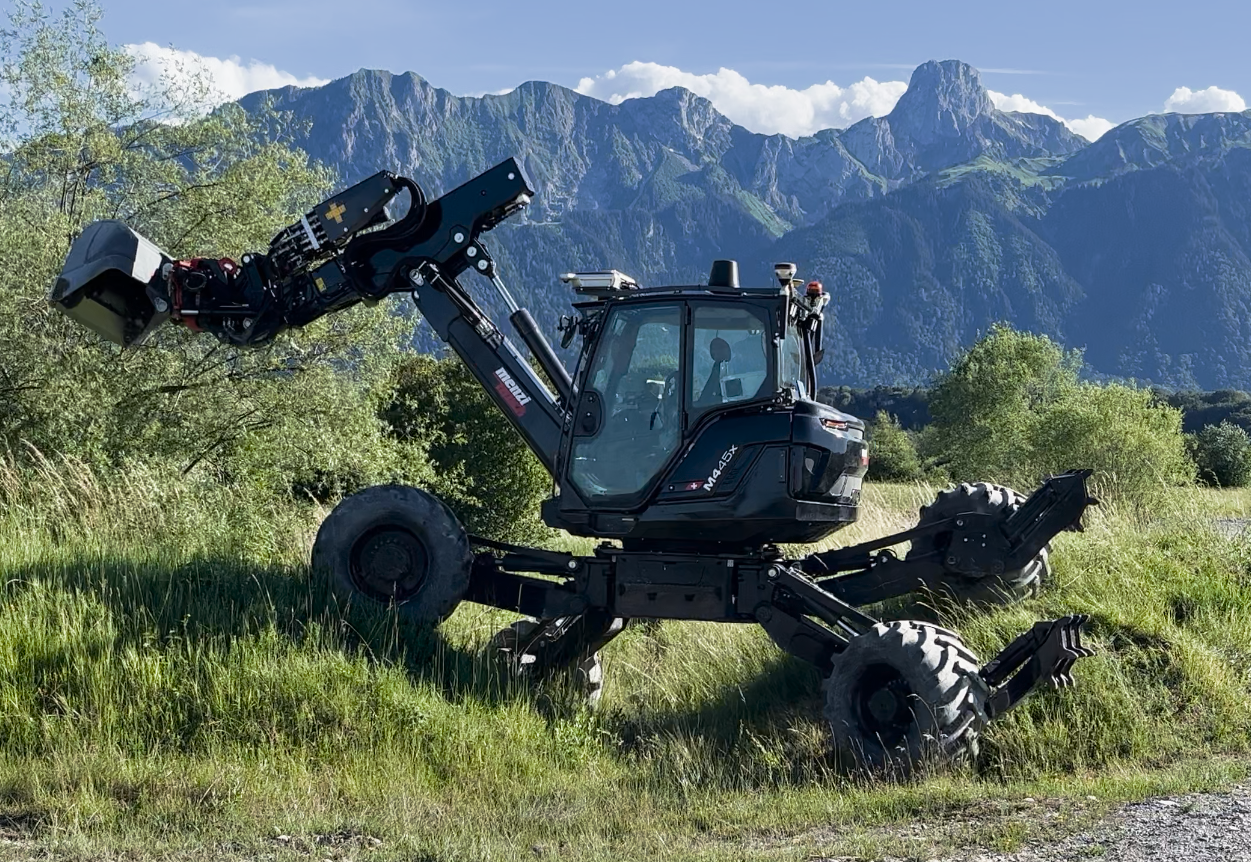}
\caption{Machine driving to the first workspace through uneven terrain. A force-adaptive chassis controller keeps the cabin level and all four wheels in ground contact.}
\label{fig:hip_balance}
\end{figure}

We transform the target embankment path from global coordinates into a local map frame and discretize it into shovel-width overlapping targets.
An offset base path, later tracked by a curvature-based driving controller, places the targets within reach of the arm.
A force-based hip-balancing controller uses the articulated legs to maintain a constant base height, horizontal base alignment and wheel contact over uneven terrain.
This support stack allows the embankment skill to operate on terrain containing rocks and boulders.
Figure~\ref{fig:hip_balance} shows the legs adapting while driving to the first workspace.

A path from the current base position to the first workspace is planned using a Reeds-Shepp-based planner from OMPL \cite{sucan2012ompl}.
To keep the legs and wheels outside the arm workspace, the target yaw $\theta$ is constrained to be tangential to the offset base path.
We add a second goal pose with yaw $\theta + \pi$ since base driving direction is ambiguous on the M4.
This allows the planner to select reverse driving instead of turning the base when it yields a shorter path.

At the target base pose, the state machine switches to \textit{MoveArmToAttackPoint}.
The shovel is first moved to an intermediate point that leaves the workspace unobstructed for the LiDAR height scan.

\begin{figure}
\centering
\includegraphics[width=0.8\linewidth]{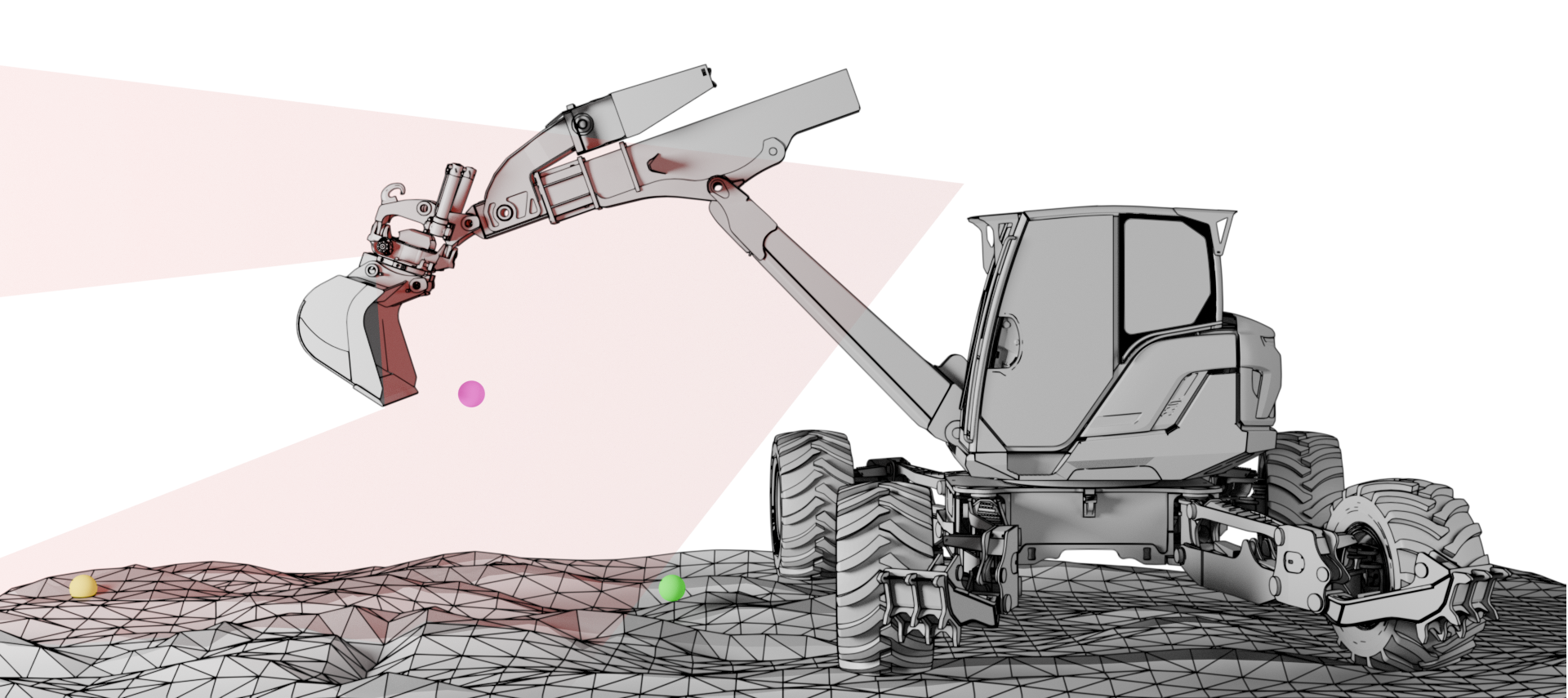}
\caption{Target (green), intermediate (pink), and attack point (yellow) for a single scoop. The area shaded in red shows the field of view of the LiDAR. A low shovel can block LiDAR visibility for the rear part of the workspace. Intermediate points are chosen to guarantee an unobstructed view of the workspace.}
\label{fig:intermPoints}
\end{figure}

To use as much available soil as possible, the attack point is placed at the far end of the workspace.
\Cref{fig:intermPoints} shows the target, intermediate, and attack points for one scoop.
Because the arm plane is not necessarily perpendicular to the embankment path, we compute the task-space target coordinate $x^{\star}$ such that the policy target coincides with the globally registered target point.
For the loose soil in this deployment, we use three scoops per target point.
After each scoop, the arm returns to the intermediate point and the height scan is updated.
A new attack point is then computed from the updated soil geometry before the next policy execution.
After all scoops at a target point are complete, the next target is selected and the cabin slews to bring it into the arm plane.
When all targets in a workspace are complete, the next workspace is selected and the state machine returns to \textit{DriveToWorkspace}.
After the final workspace, all active controllers are stopped and the state machine switches to \textit{Finished}.

\end{document}